\documentclass[letterpaper]{article} 
\usepackage[preprint]{aaai2027}  
\usepackage[hyphens]{url}  
\usepackage{graphicx} 
\usepackage{natbib}  
\usepackage{caption} 
\usepackage{algorithm}
\usepackage{algorithmic}

\usepackage{newfloat}
\usepackage{listings}
\DeclareCaptionStyle{ruled}{labelfont=normalfont,labelsep=colon,strut=off} 
\floatstyle{ruled}
\newfloat{listing}{tb}{lst}{}
\floatname{listing}{Listing}

\usepackage{amsmath}
\usepackage{amssymb}
\usepackage{bm}
\usepackage{multirow}
\usepackage{booktabs}

\title{Semantic Prior-Guided Incomplete Cross-Modal Retrieval with Noisy Labels}

\author{
Jiali Chen\textsuperscript{\rm 1},
Ruitao Pu\textsuperscript{\rm 1},
Yingke Chen\textsuperscript{\rm 3},
Dezhong Peng\textsuperscript{\rm 1,\rm 2}\corresponding
}

\affiliations{
\textsuperscript{\rm 1}College of Computer Science, Sichuan University, Chengdu 610065, China\\
\textsuperscript{\rm 2}Tianfu Jincheng Laboratory, Chengdu 610093, China\\
\textsuperscript{\rm 3}Department of Computer and Information Sciences, Northumbria University, Newcastle upon Tyne NE1 8ST, UK\\
chenjiali27.clark@gmail.com,
ruitaopu@gmail.com,
yke.chen@gmail.com,
pengdz@scu.edu.cn
}

\begin{document}

\maketitle

\begin{abstract}
Cross-modal retrieval (CMR) aims to bridge the semantic gaps between modalities by learning shared representations. Unfortunately, due to the dynamism and inconsistencies in real-world data acquisition, multimodal datasets frequently encounter severe modality incompleteness. Existing incomplete CMR methods adhere to a label-driven paradigm, relying heavily on categorical annotations as the semantic benchmark. However, label noise is difficult to avoid and often coexists with modality incompleteness. To address this challenge, we propose a novel semantic prior-guided framework for incomplete cross-modal retrieval with noisy labels, termed SPIRN. Specifically, SPIRN uses fixed semantic priors as a consistent reference for robust representation learning, noisy-label handling, and missing-modality completion. 
Based on this reference, Semantic Priors Guidance (SPG) establishes a stable category-level semantic space, Quality-aware Robust Learning (QRL) dynamically evaluates sample quality to distinguish clean samples from noisy ones and handles noise via reverse learning, Prior-guided Global Label Calibration (PGLC) rectifies corrupted annotations for refined supervision, and Confidence-driven Selective Imputation (CSI) recovers missing modalities from a cross-modal memory bank while filtering unreliable imputations with an increasing threshold.
To the best of our knowledge, this is the first framework to jointly tackle the dual challenges of modality incompleteness and label noise in the CMR community. Extensive experiments on four widely used benchmark datasets demonstrate that SPIRN outperforms 11 state-of-the-art methods in various scenarios involving both missing modalities and noisy labels.
\end{abstract}


\begin{figure}[!t]
    \centering
    \includegraphics[width=0.94\linewidth]{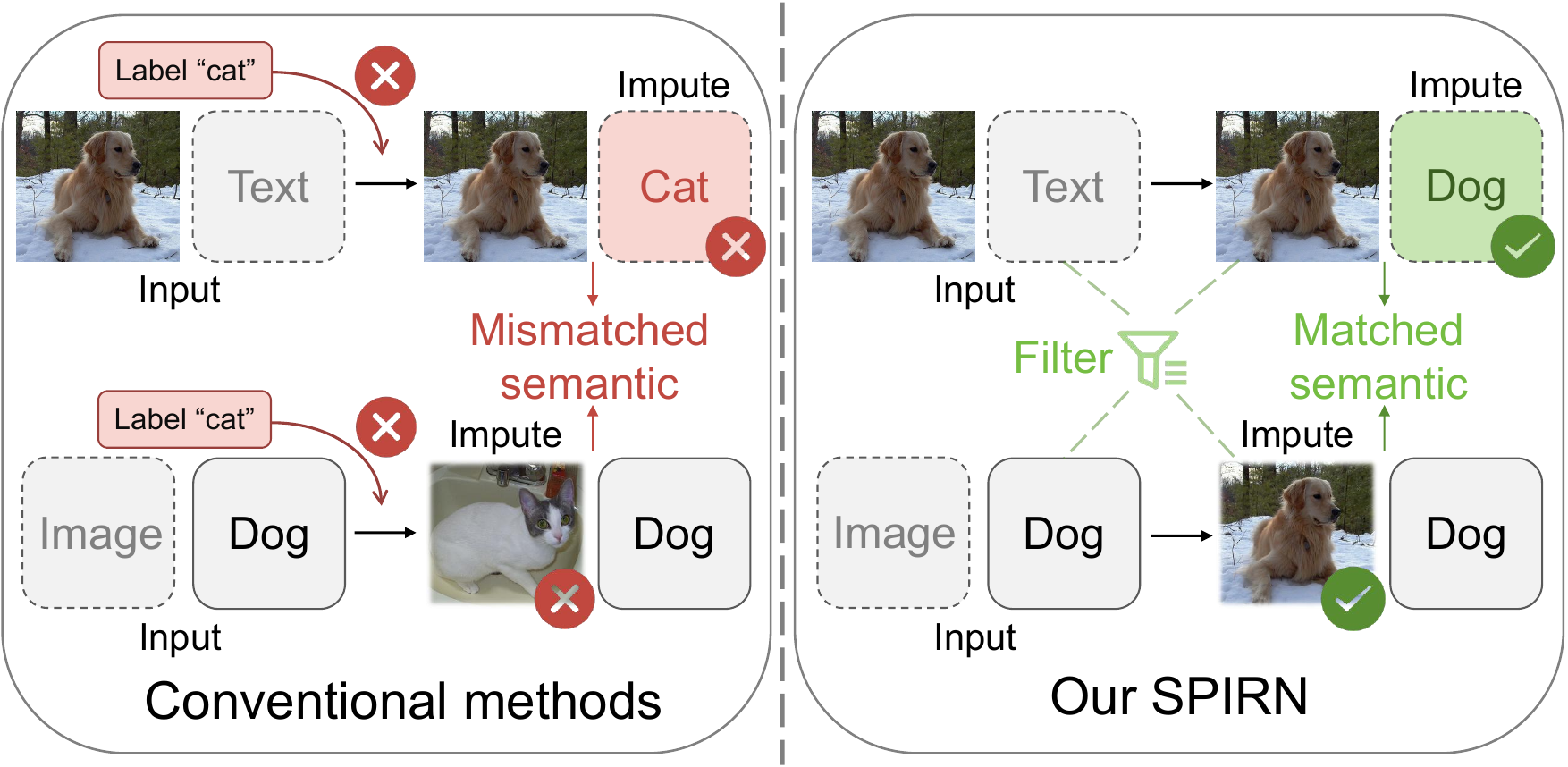}
    \caption{The motivation of SPIRN. Conventional ICMR methods rely heavily on labels and are vulnerable to noisy supervision, which may lead to semantic confusion. In contrast, SPIRN filters and exploits high-quality imputed data, thereby alleviating the impact of noisy labels.}
    \label{fig:motivation}
\end{figure}

\begin{figure*}[!t]
    \centering
    \includegraphics[width=0.85\linewidth]{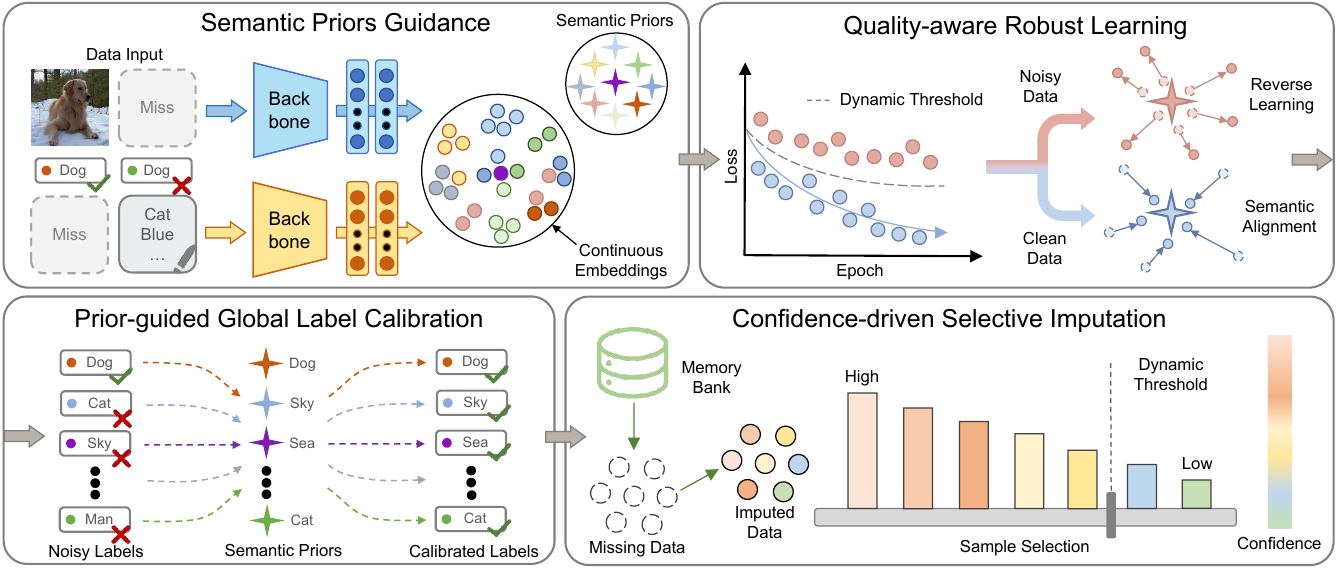}
    \caption{The framework of SPIRN consists of four core components: Semantic Priors Guidance (SPG), Quality-aware Robust Learning (QRL), Prior-guided Global Label Calibration (PGLC), and Confidence-driven Selective Imputation (CSI).}
    \label{fig:framework}
\end{figure*}

\section{Introduction}
With the increasing availability of multimodal data, cross-modal retrieval (CMR)~\cite{zhou2023state,wang2025cross} has attracted considerable attention in recent years. Its core objective is to learn discriminative cross-modal representations that capture semantic relationships between heterogeneous modalities. Existing methods typically preserve semantic consistency by exploiting label supervision~\cite{DRCL,GNN4CMR} or maintaining original semantic structures~\cite{UCCH,DGCPN}. However, most of them implicitly assume that training data are both complete and clean. In practice, this assumption rarely holds: modality completeness is often affected by communication bottlenecks~\cite{betzel2018approximate} or sensor failures~\cite{liu2020data}, while label reliability is frequently compromised by annotation errors or non-expert labeling~\cite{cheng2023intelligent}. Therefore, it is crucial to develop robust CMR methods that can learn reliable semantic representations in the presence of both incomplete multimodal data and noisy labels.

Substantial efforts have been devoted to learning cross-modal representations from incomplete multimodal data~\cite{luo2024contrastive,peng2025self,wang2020icmsc,zhang2024towards,dai2025unbiased}. For example, LNLN~\cite{zhang2024towards} improves robustness to incomplete multimodal inputs through dominant-modality correction and reconstruction, while MD$^2$N~\cite{dai2025unbiased} addresses missing-modality recovery with bidirectional generation and modality transfer. Despite their effectiveness, these methods generally assume reliable annotations. In real-world scenarios, however, high-quality labels are expensive to obtain, and alternative labeling strategies such as machine-generated annotations~\cite{kuznetsova2020open}, search-engine-based labeling~\cite{fergus2010learning}, and crowd-sourcing~\cite{welinder2010multidimensional} often introduce noisy labels~\cite{frenay2013classification}. Missing modalities weaken cross-modal correspondence and enlarge modality heterogeneity~\cite{luo2024contrastive}, while noisy labels provide misleading supervision that can drive models to overfit incorrect semantics~\cite{RSHNL}. Therefore, under coupled corruption, the key challenge is not merely to combine noise-robust learning with modality completion, but to establish a stable semantic reference that remains reliable throughout robust representation learning, noisy-label correction, and missing-modality recovery.

To tackle these limitations, this paper proposes SPIRN, a novel semantic prior-guided framework for incomplete cross-modal retrieval with noisy labels. As shown in Fig.~\ref{fig:motivation}, unlike existing methods that rely heavily on potentially corrupted labels, SPIRN uses fixed semantic priors as a consistent reference for representation learning, noisy-label correction, and missing-modality completion. As illustrated in Fig.~\ref{fig:framework}, this reference consistently supports four tightly coupled components: Semantic Priors Guidance (SPG), Quality-aware Robust Learning (QRL), Prior-guided Global Label Calibration (PGLC), and Confidence-driven Selective Imputation (CSI).
Specifically, SPG establishes a prior-guided semantic space that serves as the shared basis for subsequent learning. Building on this stable reference, QRL dynamically evaluates sample quality to distinguish clean samples from noisy ones and handles noise via reverse learning, PGLC rectifies corrupted annotations for refined supervision, and CSI recovers missing modalities from a cross-modal memory bank while filtering unreliable imputations with an increasing threshold. In this way, semantic priors are not merely auxiliary supervision, but the central mechanism that unifies robust learning and modality recovery under coupled data corruption. Overall, the main contributions are summarized as follows:
\begin{itemize}
    \item This paper proposes a novel semantic prior-guided framework for incomplete cross-modal retrieval with noisy labels, formulating modality incompleteness and label noise as coupled challenges. To the best of our knowledge, this is the first framework to jointly address these challenges in the CMR community.
    \item We propose Semantic Priors Guidance (SPG), which introduces fixed, orthogonal semantic priors as global supervisory signals, providing a stable semantic reference for learning under incomplete modalities and noisy labels.
    \item We design Quality-aware Robust Learning (QRL), Prior-guided Global Label Calibration (PGLC), and Confidence-driven Selective Imputation (CSI) to support robust sample optimization, label noise rectification, and reliable missing-modality recovery in a unified manner.
    \item Extensive experiments on four widely used benchmark datasets demonstrate that our SPIRN outperforms 11 state-of-the-art methods in various scenarios involving both missing modalities and noisy labels.
\end{itemize}

\section{Related Work}
\subsection{Incomplete Cross-Modal Retrieval}
In real-world scenarios, multimodal data are often corrupted or partially missing, which can substantially degrade retrieval performance. To address this issue, a growing body of research has investigated incomplete cross-modal retrieval~\cite{luo2024contrastive,shi2024incomplete,gong2024enhancing,gong2024cross,peng2025self}. Existing methods can be broadly grouped into two lines. One line focuses on incomplete instance-level retrieval by recovering missing modalities through cross-modal similarity modeling or nearest-neighbor construction~\cite{gong2023prototype,gong2024cross}. The other line studies incomplete category-level retrieval, which emphasizes category-level semantic consistency and is more tolerant to intra-class variations~\cite{luo2024contrastive,peng2025self,shi2024incomplete,wang2025cross}. For example, ICMR-DCT~\cite{shi2024incomplete} generates missing modalities by exploiting cross-modal mapping relationships and neighborhood correlations.

Despite their effectiveness, most existing incomplete cross-modal retrieval methods assume reliable training annotations. However, multimodal datasets are often accompanied by noisy labels in practice, which introduce misleading supervision and further complicate representation learning under modality incompleteness. Different from these works, our study considers a more challenging setting where missing modalities and label noise coexist.

\subsection{Cross-Modal Retrieval with Noisy Labels}
Learning with noisy labels has also attracted increasing attention in the cross-modal retrieval community~\cite{feng2023rono,RSHNL,liu2025neighbor,pu2024deep,okamura2023lcnme,han2025unified}. Existing methods mainly improve robustness by identifying corrupted samples or labels and applying tailored optimization strategies. For instance, LCNME~\cite{okamura2023lcnme} corrects noisy labels by exploiting memorization effects to estimate a critical change epoch, while RSHNL~\cite{RSHNL} dynamically quantifies instance learning difficulty to progressively separate clean and noisy pairs. NIRNL~\cite{liu2025neighbor} further leverages cross-modal neighborhood consensus to partition samples into different subsets for refined optimization.

However, these methods are primarily developed under the assumption of complete multimodal observations. When modality incompleteness is present, cross-modal correspondence becomes weaker, making noisy supervision more difficult to identify and correct. Therefore, a unified framework that can simultaneously address missing modalities and noisy labels is highly desirable for robust cross-modal retrieval in real-world scenarios.

\section{Method}
\subsection{Problem Formulation}

We first introduce the notations used throughout the paper. The training set is denoted by $\mathcal{O} = \mathcal{F} \cup \mathcal{V} \cup \mathcal{U}$, where $\mathcal{F} = \{(x_i^v, x_i^u, \mathbf{l}_i)\}_{i=1}^{N_f}$ denotes the complete multimodal set, $\mathcal{V} = \{(x_i^v, \mathbf{l}_i)\}_{i=N_f+1}^{N_f+N_v}$ denotes the image-only set, and $\mathcal{U} = \{(x_i^u, \mathbf{l}_i)\}_{i=N_f+N_v+1}^{N_f+N_v+N_u}$ denotes the text-only set. Here, $N = N_f + N_v + N_u$ is the total number of training instances, while $x_i^v$ and $x_i^u$ represent the image and text modalities of the $i$-th instance, respectively. The latent clean label of the $i$-th instance is denoted by $\mathbf{l}_i \in \{0,1\}^C$, where $C$ is the number of semantic categories. If the instance belongs to category $c$, then $l_{i,c}=1$; otherwise, $l_{i,c}=0$. During training, only the observed noisy label $\tilde{\mathbf{l}}_i \in \{0,1\}^C$ is available.

Modality-specific encoders are denoted by $E^*(\cdot, \Theta^*)$, where $\Theta^*$ denotes the learnable parameters and $* \in \{v,u\}$ indicates the image or text modality. For similarity measurement, the encoded features are further normalized. For each instance $x_i^*$, its raw feature $\mathbf{h}_i^*$ and final semantic representation $\mathbf{z}_i^*$ are obtained as
\begin{equation}
    \mathbf{h}_i^* = E^*(x_i^*, \Theta^*), \;
    \mathbf{z}_i^* = \frac{\mathbf{h}_i^*}{\|\mathbf{h}_i^*\|_2},
    \label{eq:1}
\end{equation}
where $\|\cdot\|_2$ denotes the $L_2$ norm, and $\mathbf{z}_i^* \in \mathbb{R}^D$ is the $L_2$-normalized continuous embedding.

\subsection{Semantic Priors Guidance}
Conventional ICMR methods \cite{NCH, luo2024contrastive, EGICH} typically operate under the idealized assumption of perfectly clean annotations, heavily relying on the observed labels as semantic guidance for cross-modal alignment and missing modality recovery. However, in real-world scenarios, missing modalities in multimodal data often coexist with inherent label noise, which is difficult for existing methods to handle effectively.

To tackle this problem, we propose a Semantic Priors Guidance (SPG) module. Instead of dynamically updating the class centers, which may be easily corrupted by the combined effect of noisy labels and missing data, we introduce a set of fixed orthogonal semantic anchors defining a stable common semantic space for aligning heterogeneous modalities. Specifically, we define a fixed orthogonal matrix $\mathbf{W} = [\mathbf{w}_1, \mathbf{w}_2, \dots, \mathbf{w}_C]^\top \in \mathbb{R}^{C \times D}$, where $\mathbf{w}_c \in \mathbb{R}^D$ denotes the semantic anchor for the $c$-th category, satisfying $\mathbf{W} \mathbf{W}^\top = \mathbf{I}$. Given the $L_2$-normalized embeddings $\mathbf{z}_i^v$ and $\mathbf{z}_i^u$, we compute the category probability distribution using a scaled softmax function:
\begin{equation}
    p(c \mid x_i^*) =
    \frac{\exp\!\left((\mathbf{z}_i^*)^\top \mathbf{w}_c / \tau\right)}
    {\sum_{j=1}^{C} \exp\!\left((\mathbf{z}_i^*)^\top \mathbf{w}_j / \tau\right)},
    \; * \in \{v,u\},
    \label{eq:2}
\end{equation}
where $\mathbf{z}_i^*$ denotes the semantic embedding of modality $*$, and $\tau$ is a temperature parameter that controls the sharpness of the probability distribution. Based on the semantic priors, we define the SPG loss as the cross-entropy loss between the predicted probabilities and the observed noisy labels on the complete dataset $\mathcal{F}$:
\begin{equation}
    \mathcal{L}_{\mathrm{spg}} =
    -\frac{1}{N_f} \sum_{i=1}^{N_f} \sum_{c=1}^{C} \tilde{l}_{i,c}
    \left( \log p(c \mid x_i^v) + \log p(c \mid x_i^u) \right).
    \label{eq:3}
\end{equation}

\begin{table*}[!t]
\centering
{\small
\setlength{\tabcolsep}{4.5pt}
\begin{tabular}{c|l|cccccccc|cccccccc}
\hline
\multirow{3}{*}{Missing} & \multicolumn{1}{c|}{\multirow{3}{*}{Method}} & \multicolumn{8}{c|}{Wikipedia} & \multicolumn{8}{c}{XMedia} \\ \cline{3-18} 
 & \multicolumn{1}{c|}{} & \multicolumn{2}{c|}{0.2} & \multicolumn{2}{c|}{0.4} & \multicolumn{2}{c|}{0.6} & \multicolumn{2}{c|}{0.8} & \multicolumn{2}{c|}{0.2} & \multicolumn{2}{c|}{0.4} & \multicolumn{2}{c|}{0.6} & \multicolumn{2}{c}{0.8} \\ \cline{3-18} 
 & \multicolumn{1}{c|}{} & I2T & \multicolumn{1}{c|}{T2I} & I2T & \multicolumn{1}{c|}{T2I} & I2T & \multicolumn{1}{c|}{T2I} & I2T & T2I & I2T & \multicolumn{1}{c|}{T2I} & I2T & \multicolumn{1}{c|}{T2I} & I2T & \multicolumn{1}{c|}{T2I} & I2T & T2I \\ \hline
\multirow{12}{*}{Easy} & ALGCN & 37.2 & \multicolumn{1}{c|}{34.4} & 28.6 & \multicolumn{1}{c|}{27.6} & 19.4 & \multicolumn{1}{c|}{19.7} & 14.0 & 14.7 & 86.3 & \multicolumn{1}{c|}{85.6} & 76.0 & \multicolumn{1}{c|}{73.2} & 51.1 & \multicolumn{1}{c|}{46.0} & 13.5 & 11.4 \\
 & UCCH & 35.6 & \multicolumn{1}{c|}{34.3} & 35.6 & \multicolumn{1}{c|}{34.3} & 35.6 & \multicolumn{1}{c|}{34.3} & 35.6 & 34.3 & 73.6 & \multicolumn{1}{c|}{73.5} & 73.6 & \multicolumn{1}{c|}{73.5} & 73.6 & \multicolumn{1}{c|}{73.5} & 73.6 & 73.5 \\
 & GNN4CMR & 44.4 & \multicolumn{1}{c|}{40.5} & 40.7 & \multicolumn{1}{c|}{37.2} & 31.6 & \multicolumn{1}{c|}{29.1} & 19.9 & 19.6 & 85.1 & \multicolumn{1}{c|}{84.8} & 79.3 & \multicolumn{1}{c|}{80.3} & 70.5 & \multicolumn{1}{c|}{71.3} & 42.9 & 42.7 \\
 & RONO & {\underline{49.4}} & \multicolumn{1}{c|}{\textbf{45.2}} & {\underline{48.8}} & \multicolumn{1}{c|}{{\underline{45.1}}} & {\underline{47.4}} & \multicolumn{1}{c|}{{\underline{43.3}}} & {\underline{37.3}} & {\underline{35.4}} & {\underline{90.3}} & \multicolumn{1}{c|}{{\underline{90.7}}} & {\underline{89.3}} & \multicolumn{1}{c|}{{\underline{89.8}}} & 86.3 & \multicolumn{1}{c|}{86.8} & 81.6 & 81.0 \\
 & CIRH & 16.8 & \multicolumn{1}{c|}{16.9} & 16.8 & \multicolumn{1}{c|}{16.9} & 16.8 & \multicolumn{1}{c|}{16.9} & 16.8 & 16.9 & 81.0 & \multicolumn{1}{c|}{80.7} & 81.0 & \multicolumn{1}{c|}{80.7} & 81.0 & \multicolumn{1}{c|}{80.7} & 81.0 & 80.7 \\
 & NCH & 18.1 & \multicolumn{1}{c|}{17.8} & 14.5 & \multicolumn{1}{c|}{14.8} & 13.9 & \multicolumn{1}{c|}{14.4} & 12.9 & 13.2 & 11.7 & \multicolumn{1}{c|}{9.3} & 9.2 & \multicolumn{1}{c|}{8.3} & 6.6 & \multicolumn{1}{c|}{7.7} & 6.4 & 7.0 \\
 & NRCH & 40.4 & \multicolumn{1}{c|}{38.6} & 37.5 & \multicolumn{1}{c|}{36.9} & 30.4 & \multicolumn{1}{c|}{29.7} & 30.2 & 29.8 & 88.1 & \multicolumn{1}{c|}{89.2} & 85.2 & \multicolumn{1}{c|}{85.2} & 84.9 & \multicolumn{1}{c|}{86.2} & 82.7 & 83.3 \\
 & CICH & 14.9 & \multicolumn{1}{c|}{12.7} & 13.1 & \multicolumn{1}{c|}{12.7} & 12.5 & \multicolumn{1}{c|}{12.6} & 12.6 & 12.6 & 8.5 & \multicolumn{1}{c|}{8.3} & 6.8 & \multicolumn{1}{c|}{7.4} & 6.1 & \multicolumn{1}{c|}{6.5} & 6.2 & 6.6 \\
 & RSHNL & 48.2 & \multicolumn{1}{c|}{43.9} & 44.4 & \multicolumn{1}{c|}{40.9} & 35.6 & \multicolumn{1}{c|}{33.1} & 24.2 & 23.2 & 90.0 & \multicolumn{1}{c|}{90.3} & 88.9 & \multicolumn{1}{c|}{89.6} & {\underline{88.9}} & \multicolumn{1}{c|}{{\underline{89.0}}} & {\underline{87.1}} & {\underline{87.7}} \\
 & DHRL & 47.4 & \multicolumn{1}{c|}{43.9} & 43.8 & \multicolumn{1}{c|}{41.8} & 33.0 & \multicolumn{1}{c|}{32.2} & 21.7 & 23.3 & \textbf{90.6} & \multicolumn{1}{c|}{90.5} & 85.8 & \multicolumn{1}{c|}{84.4} & 71.6 & \multicolumn{1}{c|}{69.4} & 40.2 & 41.3 \\
 & EGICH & 45.6 & \multicolumn{1}{c|}{40.1} & 35.7 & \multicolumn{1}{c|}{30.7} & 24.8 & \multicolumn{1}{c|}{23.7} & 17.2 & 16.9 & 85.4 & \multicolumn{1}{c|}{84.6} & 75.0 & \multicolumn{1}{c|}{74.9} & 58.9 & \multicolumn{1}{c|}{58.5} & 34.0 & 33.7 \\
 & SPIRN & \textbf{49.7} & \multicolumn{1}{c|}{{\underline{45.1}}} & \textbf{49.9} & \multicolumn{1}{c|}{\textbf{45.9}} & \textbf{47.6} & \multicolumn{1}{c|}{\textbf{43.9}} & \textbf{42.1} & \textbf{39.1} & \textbf{90.6} & \multicolumn{1}{c|}{\textbf{91.6}} & \textbf{90.8} & \multicolumn{1}{c|}{\textbf{91.6}} & \textbf{91.1} & \multicolumn{1}{c|}{\textbf{91.4}} & \textbf{90.8} & \textbf{91.6} \\ \hline
\multirow{12}{*}{Medium} & ALGCN & 31.4 & \multicolumn{1}{c|}{30.0} & 24.0 & \multicolumn{1}{c|}{24.4} & 15.8 & \multicolumn{1}{c|}{16.3} & 13.3 & 14.4 & 82.5 & \multicolumn{1}{c|}{80.2} & 70.8 & \multicolumn{1}{c|}{62.6} & 40.5 & \multicolumn{1}{c|}{29.0} & 11.6 & 8.8 \\
 & UCCH & 29.4 & \multicolumn{1}{c|}{28.6} & 29.4 & \multicolumn{1}{c|}{28.6} & 29.4 & \multicolumn{1}{c|}{28.6} & {\underline{29.4}} & {\underline{28.6}} & 72.8 & \multicolumn{1}{c|}{74.4} & 72.8 & \multicolumn{1}{c|}{74.4} & 72.8 & \multicolumn{1}{c|}{74.4} & 72.8 & 74.4 \\
 & GNN4CMR & 41.7 & \multicolumn{1}{c|}{38.4} & 37.5 & \multicolumn{1}{c|}{34.1} & 28.0 & \multicolumn{1}{c|}{25.8} & 19.7 & 19.4 & 81.9 & \multicolumn{1}{c|}{82.3} & 75.9 & \multicolumn{1}{c|}{76.0} & 64.6 & \multicolumn{1}{c|}{65.0} & 39.7 & 39.4 \\
 & RONO & 44.6 & \multicolumn{1}{c|}{40.6} & 39.1 & \multicolumn{1}{c|}{36.4} & {\underline{36.1}} & \multicolumn{1}{c|}{{\underline{34.9}}} & 27.2 & 26.3 & 88.2 & \multicolumn{1}{c|}{{\underline{89.6}}} & 85.9 & \multicolumn{1}{c|}{87.1} & 83.0 & \multicolumn{1}{c|}{82.6} & 76.7 & 75.2 \\
 & CIRH & 14.8 & \multicolumn{1}{c|}{15.0} & 14.8 & \multicolumn{1}{c|}{15.0} & 14.8 & \multicolumn{1}{c|}{15.0} & 14.8 & 15.0 & 62.9 & \multicolumn{1}{c|}{60.2} & 62.9 & \multicolumn{1}{c|}{60.2} & 62.9 & \multicolumn{1}{c|}{60.2} & 62.9 & 60.2 \\
 & NCH & 16.8 & \multicolumn{1}{c|}{18.0} & 14.6 & \multicolumn{1}{c|}{15.2} & 13.1 & \multicolumn{1}{c|}{13.2} & 13.0 & 13.3 & 10.0 & \multicolumn{1}{c|}{9.3} & 8.4 & \multicolumn{1}{c|}{8.2} & 7.2 & \multicolumn{1}{c|}{8.0} & 6.4 & 7.6 \\
 & NRCH & 35.2 & \multicolumn{1}{c|}{34.5} & 34.0 & \multicolumn{1}{c|}{32.5} & 25.1 & \multicolumn{1}{c|}{25.3} & 26.2 & 25.4 & 83.6 & \multicolumn{1}{c|}{84.7} & 81.3 & \multicolumn{1}{c|}{81.7} & 81.0 & \multicolumn{1}{c|}{81.5} & 78.7 & 78.8 \\
 & CICH & 21.3 & \multicolumn{1}{c|}{15.0} & 12.7 & \multicolumn{1}{c|}{13.3} & 12.4 & \multicolumn{1}{c|}{12.6} & 12.6 & 12.6 & 6.7 & \multicolumn{1}{c|}{7.1} & 6.2 & \multicolumn{1}{c|}{7.0} & 6.1 & \multicolumn{1}{c|}{6.8} & 6.1 & 6.4 \\
 & RSHNL & {\underline{45.7}} & \multicolumn{1}{c|}{40.0} & {\underline{41.9}} & \multicolumn{1}{c|}{{\underline{38.6}}} & 31.4 & \multicolumn{1}{c|}{28.9} & 21.0 & 20.1 & {\underline{88.8}} & \multicolumn{1}{c|}{88.9} & {\underline{86.8}} & \multicolumn{1}{c|}{{\underline{87.5}}} & {\underline{85.9}} & \multicolumn{1}{c|}{{\underline{85.4}}} & {\underline{84.0}} & {\underline{84.0}} \\
 & DHRL & 43.9 & \multicolumn{1}{c|}{40.3} & 39.1 & \multicolumn{1}{c|}{36.0} & 27.8 & \multicolumn{1}{c|}{26.8} & 19.7 & 19.8 & 86.3 & \multicolumn{1}{c|}{86.2} & 79.9 & \multicolumn{1}{c|}{79.0} & 66.1 & \multicolumn{1}{c|}{61.2} & 26.5 & 25.2 \\
 & EGICH & 45.4 & \multicolumn{1}{c|}{{\underline{40.9}}} & 33.7 & \multicolumn{1}{c|}{29.8} & 22.3 & \multicolumn{1}{c|}{21.4} & 16.2 & 15.9 & 83.9 & \multicolumn{1}{c|}{84.8} & 68.7 & \multicolumn{1}{c|}{68.2} & 47.0 & \multicolumn{1}{c|}{46.5} & 23.9 & 24.5 \\
 & SPIRN & \textbf{47.4} & \multicolumn{1}{c|}{\textbf{42.7}} & \textbf{46.7} & \multicolumn{1}{c|}{\textbf{42.3}} & \textbf{45.6} & \multicolumn{1}{c|}{\textbf{41.4}} & \textbf{39.9} & \textbf{37.0} & \textbf{90.2} & \multicolumn{1}{c|}{\textbf{91.2}} & \textbf{89.8} & \multicolumn{1}{c|}{\textbf{91.1}} & \textbf{89.7} & \multicolumn{1}{c|}{\textbf{90.0}} & \textbf{89.4} & \textbf{90.2} \\ \hline
\multirow{12}{*}{Hard} & ALGCN & 24.6 & \multicolumn{1}{c|}{24.3} & 19.8 & \multicolumn{1}{c|}{20.3} & 15.1 & \multicolumn{1}{c|}{15.5} & 13.3 & 13.8 & 67.0 & \multicolumn{1}{c|}{41.9} & 47.2 & \multicolumn{1}{c|}{22.2} & 21.2 & \multicolumn{1}{c|}{10.9} & 10.3 & 8.0 \\
 & UCCH & 22.0 & \multicolumn{1}{c|}{21.6} & 22.0 & \multicolumn{1}{c|}{21.6} & 22.0 & \multicolumn{1}{c|}{21.6} & {\underline{22.0}} & {\underline{21.6}} & 62.7 & \multicolumn{1}{c|}{66.0} & 62.7 & \multicolumn{1}{c|}{66.0} & 62.7 & \multicolumn{1}{c|}{66.0} & 62.7 & 66.0 \\
 & GNN4CMR & 34.2 & \multicolumn{1}{c|}{31.7} & 27.1 & \multicolumn{1}{c|}{24.9} & {\underline{23.1}} & \multicolumn{1}{c|}{22.2} & 17.1 & 17.3 & 72.7 & \multicolumn{1}{c|}{74.5} & 63.3 & \multicolumn{1}{c|}{65.9} & 50.4 & \multicolumn{1}{c|}{51.5} & 34.1 & 33.6 \\
 & RONO & 23.3 & \multicolumn{1}{c|}{22.8} & 20.9 & \multicolumn{1}{c|}{21.1} & 20.3 & \multicolumn{1}{c|}{20.1} & 18.3 & 17.9 & {\underline{84.1}} & \multicolumn{1}{c|}{{\underline{84.5}}} & 75.7 & \multicolumn{1}{c|}{73.9} & 67.2 & \multicolumn{1}{c|}{63.8} & 52.6 & 50.3 \\
 & CIRH & 13.4 & \multicolumn{1}{c|}{13.9} & 13.4 & \multicolumn{1}{c|}{13.9} & 13.4 & \multicolumn{1}{c|}{13.9} & 13.4 & 13.9 & 52.1 & \multicolumn{1}{c|}{48.6} & 52.1 & \multicolumn{1}{c|}{48.6} & 52.1 & \multicolumn{1}{c|}{48.6} & 52.1 & 48.6 \\
 & NCH & 16.3 & \multicolumn{1}{c|}{16.6} & 14.0 & \multicolumn{1}{c|}{13.8} & 13.7 & \multicolumn{1}{c|}{13.6} & 13.8 & 13.8 & 10.0 & \multicolumn{1}{c|}{8.0} & 7.4 & \multicolumn{1}{c|}{7.9} & 6.8 & \multicolumn{1}{c|}{8.5} & 6.1 & 7.4 \\
 & NRCH & 25.5 & \multicolumn{1}{c|}{24.8} & 21.8 & \multicolumn{1}{c|}{21.4} & 20.5 & \multicolumn{1}{c|}{20.0} & 19.4 & 18.9 & 74.8 & \multicolumn{1}{c|}{75.0} & 71.2 & \multicolumn{1}{c|}{72.0} & 64.6 & \multicolumn{1}{c|}{65.2} & 63.9 & 64.4 \\
 & CICH & 17.3 & \multicolumn{1}{c|}{19.9} & 13.9 & \multicolumn{1}{c|}{13.6} & 12.6 & \multicolumn{1}{c|}{15.4} & 12.6 & 12.5 & 6.3 & \multicolumn{1}{c|}{6.9} & 6.3 & \multicolumn{1}{c|}{7.8} & 6.1 & \multicolumn{1}{c|}{7.2} & 6.3 & 6.9 \\
 & RSHNL & 36.6 & \multicolumn{1}{c|}{33.8} & 31.5 & \multicolumn{1}{c|}{28.7} & 22.0 & \multicolumn{1}{c|}{21.7} & 16.0 & 16.2 & 83.2 & \multicolumn{1}{c|}{82.4} & {\underline{81.4}} & \multicolumn{1}{c|}{{\underline{81.9}}} & {\underline{75.1}} & \multicolumn{1}{c|}{{\underline{74.2}}} & {\underline{68.7}} & {\underline{67.3}} \\
 & DHRL & 33.9 & \multicolumn{1}{c|}{31.5} & 27.1 & \multicolumn{1}{c|}{27.1} & {\underline{23.1}} & \multicolumn{1}{c|}{{\underline{22.7}}} & 15.7 & 16.1 & 76.7 & \multicolumn{1}{c|}{76.5} & 63.1 & \multicolumn{1}{c|}{63.0} & 36.9 & \multicolumn{1}{c|}{36.8} & 20.3 & 18.9 \\
 & EGICH & {\underline{43.5}} & \multicolumn{1}{c|}{{\underline{39.0}}} & {\underline{32.9}} & \multicolumn{1}{c|}{{\underline{29.1}}} & 21.2 & \multicolumn{1}{c|}{19.6} & 13.7 & 13.7 & 79.6 & \multicolumn{1}{c|}{80.7} & 64.8 & \multicolumn{1}{c|}{63.7} & 38.4 & \multicolumn{1}{c|}{38.5} & 17.1 & 17.7 \\
 & SPIRN & \textbf{46.3} & \multicolumn{1}{c|}{\textbf{40.4}} & \textbf{45.4} & \multicolumn{1}{c|}{\textbf{40.0}} & \textbf{44.5} & \multicolumn{1}{c|}{\textbf{39.3}} & \textbf{33.8} & \textbf{30.6} & \textbf{88.5} & \multicolumn{1}{c|}{\textbf{89.3}} & \textbf{86.7} & \multicolumn{1}{c|}{\textbf{87.3}} & \textbf{88.2} & \multicolumn{1}{c|}{\textbf{89.1}} & \textbf{84.3} & \textbf{85.0} \\ \hline
\end{tabular}
}
\caption{Comparison of mAP (\%) on Wikipedia and XMedia under different completeness and noise settings.}
\label{tab:wiki_xmedia_comparison_results}
\end{table*}

\begin{table*}[!t]
\centering
{\small
\setlength{\tabcolsep}{4.5pt}      
\begin{tabular}{c|l|cccccccc|cccccccc}
\hline
\multirow{3}{*}{Missing} & \multicolumn{1}{c|}{\multirow{3}{*}{Method}} & \multicolumn{8}{c|}{INRIA-Websearch} & \multicolumn{8}{c}{XMediaNet} \\ \cline{3-18} 
 & \multicolumn{1}{c|}{} & \multicolumn{2}{c|}{0.2} & \multicolumn{2}{c|}{0.4} & \multicolumn{2}{c|}{0.6} & \multicolumn{2}{c|}{0.8} & \multicolumn{2}{c|}{0.2} & \multicolumn{2}{c|}{0.4} & \multicolumn{2}{c|}{0.6} & \multicolumn{2}{c}{0.8} \\ \cline{3-18} 
 & \multicolumn{1}{c|}{} & I2T & \multicolumn{1}{c|}{T2I} & I2T & \multicolumn{1}{c|}{T2I} & I2T & \multicolumn{1}{c|}{T2I} & I2T & T2I & I2T & \multicolumn{1}{c|}{T2I} & I2T & \multicolumn{1}{c|}{T2I} & I2T & \multicolumn{1}{c|}{T2I} & I2T & T2I \\ \hline
\multirow{12}{*}{Easy} & ALGCN & 32.3 & \multicolumn{1}{c|}{30.3} & 19.5 & \multicolumn{1}{c|}{16.9} & 6.4 & \multicolumn{1}{c|}{4.1} & 2.0 & 1.7 & 21.4 & \multicolumn{1}{c|}{26.1} & 8.0 & \multicolumn{1}{c|}{10.6} & 1.7 & \multicolumn{1}{c|}{2.2} & 0.8 & 1.1 \\
 & UCCH & 35.6 & \multicolumn{1}{c|}{35.5} & 35.6 & \multicolumn{1}{c|}{35.5} & 35.6 & \multicolumn{1}{c|}{35.5} & 35.6 & 35.5 & 24.1 & \multicolumn{1}{c|}{25.8} & 24.1 & \multicolumn{1}{c|}{25.8} & 24.1 & \multicolumn{1}{c|}{25.8} & 24.1 & 25.8 \\
 & GNN4CMR & 45.5 & \multicolumn{1}{c|}{45.2} & 40.2 & \multicolumn{1}{c|}{38.7} & 34.3 & \multicolumn{1}{c|}{32.7} & 28.5 & 26.8 & {\underline{52.0}} & \multicolumn{1}{c|}{52.4} & 46.6 & \multicolumn{1}{c|}{47.2} & 41.2 & \multicolumn{1}{c|}{41.8} & 36.6 & 37.4 \\
 & RONO & 44.4 & \multicolumn{1}{c|}{44.9} & 43.0 & \multicolumn{1}{c|}{43.0} & 40.4 & \multicolumn{1}{c|}{40.1} & 38.8 & 37.7 & 35.1 & \multicolumn{1}{c|}{33.7} & 37.0 & \multicolumn{1}{c|}{34.8} & 37.0 & \multicolumn{1}{c|}{35.8} & 30.3 & 30.5 \\
 & CIRH & 32.4 & \multicolumn{1}{c|}{32.6} & 32.4 & \multicolumn{1}{c|}{32.6} & 32.4 & \multicolumn{1}{c|}{32.6} & 32.4 & 32.6 & 14.1 & \multicolumn{1}{c|}{13.4} & 14.1 & \multicolumn{1}{c|}{13.4} & 14.1 & \multicolumn{1}{c|}{13.4} & 14.1 & 13.4 \\
 & NCH & 3.9 & \multicolumn{1}{c|}{7.0} & 3.7 & \multicolumn{1}{c|}{5.3} & 2.1 & \multicolumn{1}{c|}{3.7} & 2.3 & 3.1 & 0.9 & \multicolumn{1}{c|}{1.2} & 0.8 & \multicolumn{1}{c|}{0.9} & 0.7 & \multicolumn{1}{c|}{0.9} & 0.7 & 0.8 \\
 & NRCH & 38.8 & \multicolumn{1}{c|}{39.8} & 38.6 & \multicolumn{1}{c|}{40.1} & 38.3 & \multicolumn{1}{c|}{39.1} & 37.9 & 39.1 & 17.1 & \multicolumn{1}{c|}{19.7} & 17.2 & \multicolumn{1}{c|}{20.2} & 17.1 & \multicolumn{1}{c|}{20.0} & 17.3 & 20.0 \\
 & CICH & 3.7 & \multicolumn{1}{c|}{3.5} & 4.0 & \multicolumn{1}{c|}{4.0} & 4.0 & \multicolumn{1}{c|}{3.8} & 4.0 & 4.0 & 1.9 & \multicolumn{1}{c|}{2.2} & 1.6 & \multicolumn{1}{c|}{1.5} & 1.2 & \multicolumn{1}{c|}{1.1} & 0.9 & 0.8 \\
 & RSHNL & {\underline{48.6}} & \multicolumn{1}{c|}{49.8} & {\underline{46.7}} & \multicolumn{1}{c|}{{\underline{47.8}}} & {\underline{43.4}} & \multicolumn{1}{c|}{{\underline{44.4}}} & {\underline{40.8}} & {\underline{41.6}} & 44.6 & \multicolumn{1}{c|}{45.8} & 44.8 & \multicolumn{1}{c|}{46.1} & 43.3 & \multicolumn{1}{c|}{45.2} & 41.9 & 43.4 \\
 & DHRL & 32.6 & \multicolumn{1}{c|}{33.2} & 22.0 & \multicolumn{1}{c|}{22.4} & 11.2 & \multicolumn{1}{c|}{12.5} & 4.5 & 4.7 & 18.1 & \multicolumn{1}{c|}{20.1} & 9.6 & \multicolumn{1}{c|}{13.5} & 7.9 & \multicolumn{1}{c|}{8.7} & 2.2 & 2.9 \\
 & EGICH & 48.3 & \multicolumn{1}{c|}{{\underline{49.9}}} & 44.0 & \multicolumn{1}{c|}{45.4} & 39.2 & \multicolumn{1}{c|}{40.4} & 35.0 & 35.6 & 50.7 & \multicolumn{1}{c|}{{\underline{52.5}}} & {\underline{49.5}} & \multicolumn{1}{c|}{{\underline{50.6}}} & {\underline{47.7}} & \multicolumn{1}{c|}{{\underline{48.2}}} & {\underline{46.9}} & {\underline{48.0}} \\
 & SPIRN & \textbf{53.7} & \multicolumn{1}{c|}{\textbf{55.0}} & \textbf{53.0} & \multicolumn{1}{c|}{\textbf{54.3}} & \textbf{51.2} & \multicolumn{1}{c|}{\textbf{52.6}} & \textbf{47.7} & \textbf{48.3} & \textbf{59.4} & \multicolumn{1}{c|}{\textbf{59.9}} & \textbf{59.6} & \multicolumn{1}{c|}{\textbf{60.0}} & \textbf{59.4} & \multicolumn{1}{c|}{\textbf{60.2}} & \textbf{56.9} & \textbf{57.2} \\ \hline
\multirow{12}{*}{Medium} & ALGCN & 27.5 & \multicolumn{1}{c|}{23.6} & 16.2 & \multicolumn{1}{c|}{12.0} & 5.5 & \multicolumn{1}{c|}{3.3} & 1.9 & 1.5 & 6.7 & \multicolumn{1}{c|}{9.1} & 2.5 & \multicolumn{1}{c|}{3.4} & 0.9 & \multicolumn{1}{c|}{1.3} & 0.8 & 1.1 \\
 & UCCH & 31.8 & \multicolumn{1}{c|}{31.3} & 31.8 & \multicolumn{1}{c|}{31.3} & 31.8 & \multicolumn{1}{c|}{31.3} & 31.8 & 31.3 & 17.0 & \multicolumn{1}{c|}{19.3} & 17.0 & \multicolumn{1}{c|}{19.3} & 17.0 & \multicolumn{1}{c|}{19.3} & 17.0 & 19.3 \\
 & GNN4CMR & 39.8 & \multicolumn{1}{c|}{39.7} & 33.9 & \multicolumn{1}{c|}{32.8} & 28.3 & \multicolumn{1}{c|}{27.0} & 23.5 & 21.7 & {\underline{47.6}} & \multicolumn{1}{c|}{{\underline{48.3}}} & 40.6 & \multicolumn{1}{c|}{41.8} & 33.8 & \multicolumn{1}{c|}{34.6} & 29.6 & 30.5 \\
 & RONO & 38.2 & \multicolumn{1}{c|}{38.1} & 35.0 & \multicolumn{1}{c|}{34.1} & 31.5 & \multicolumn{1}{c|}{30.7} & 30.9 & 29.7 & 30.9 & \multicolumn{1}{c|}{29.4} & 30.8 & \multicolumn{1}{c|}{28.6} & 30.0 & \multicolumn{1}{c|}{28.3} & 23.8 & 23.8 \\
 & CIRH & 27.9 & \multicolumn{1}{c|}{27.6} & 27.9 & \multicolumn{1}{c|}{27.6} & 27.9 & \multicolumn{1}{c|}{27.6} & 27.9 & 27.6 & 11.5 & \multicolumn{1}{c|}{11.2} & 11.5 & \multicolumn{1}{c|}{11.2} & 11.5 & \multicolumn{1}{c|}{11.2} & 11.5 & 11.2 \\
 & NCH & 2.8 & \multicolumn{1}{c|}{4.1} & 3.2 & \multicolumn{1}{c|}{4.0} & 2.1 & \multicolumn{1}{c|}{2.4} & 1.9 & 2.1 & 0.9 & \multicolumn{1}{c|}{1.4} & 0.8 & \multicolumn{1}{c|}{1.1} & 0.8 & \multicolumn{1}{c|}{0.9} & 0.8 & 0.9 \\
 & NRCH & 32.7 & \multicolumn{1}{c|}{33.5} & 32.5 & \multicolumn{1}{c|}{33.4} & 32.2 & \multicolumn{1}{c|}{32.9} & 31.7 & 32.2 & 15.0 & \multicolumn{1}{c|}{17.4} & 14.7 & \multicolumn{1}{c|}{17.4} & 14.9 & \multicolumn{1}{c|}{17.1} & 14.4 & 16.6 \\
 & CICH & 4.0 & \multicolumn{1}{c|}{4.0} & 3.9 & \multicolumn{1}{c|}{4.0} & 4.0 & \multicolumn{1}{c|}{4.0} & 4.0 & 4.0 & 1.7 & \multicolumn{1}{c|}{2.2} & 1.5 & \multicolumn{1}{c|}{1.9} & 1.2 & \multicolumn{1}{c|}{1.1} & 0.9 & 0.7 \\
 & RSHNL & 42.6 & \multicolumn{1}{c|}{43.8} & {\underline{40.2}} & \multicolumn{1}{c|}{{\underline{41.1}}} & {\underline{38.1}} & \multicolumn{1}{c|}{{\underline{38.8}}} & {\underline{35.6}} & {\underline{36.1}} & 38.6 & \multicolumn{1}{c|}{41.0} & 37.3 & \multicolumn{1}{c|}{39.5} & 35.6 & \multicolumn{1}{c|}{37.1} & 33.5 & 35.3 \\
 & DHRL & 26.6 & \multicolumn{1}{c|}{26.8} & 19.4 & \multicolumn{1}{c|}{18.7} & 8.8 & \multicolumn{1}{c|}{8.8} & 3.0 & 3.7 & 14.4 & \multicolumn{1}{c|}{17.3} & 8.2 & \multicolumn{1}{c|}{11.4} & 4.8 & \multicolumn{1}{c|}{6.8} & 1.3 & 2.0 \\
 & EGICH & {\underline{43.7}} & \multicolumn{1}{c|}{{\underline{44.7}}} & 36.3 & \multicolumn{1}{c|}{37.1} & 30.8 & \multicolumn{1}{c|}{31.4} & 25.2 & 25.5 & 44.9 & \multicolumn{1}{c|}{46.8} & {\underline{41.3}} & \multicolumn{1}{c|}{{\underline{43.5}}} & {\underline{37.3}} & \multicolumn{1}{c|}{{\underline{38.7}}} & {\underline{35.1}} & {\underline{36.9}} \\
 & SPIRN & \textbf{50.6} & \multicolumn{1}{c|}{\textbf{51.7}} & \textbf{49.0} & \multicolumn{1}{c|}{\textbf{50.1}} & \textbf{46.0} & \multicolumn{1}{c|}{\textbf{47.1}} & \textbf{39.1} & \textbf{40.0} & \textbf{56.0} & \multicolumn{1}{c|}{\textbf{56.8}} & \textbf{55.9} & \multicolumn{1}{c|}{\textbf{56.2}} & \textbf{55.8} & \multicolumn{1}{c|}{\textbf{55.8}} & \textbf{53.7} & \textbf{53.2} \\ \hline
\multirow{12}{*}{Hard} & ALGCN & 4.0 & \multicolumn{1}{c|}{2.6} & 2.8 & \multicolumn{1}{c|}{1.8} & 1.7 & \multicolumn{1}{c|}{1.6} & 1.4 & 1.4 & 1.0 & \multicolumn{1}{c|}{1.5} & 0.9 & \multicolumn{1}{c|}{1.3} & 0.8 & \multicolumn{1}{c|}{1.1} & 0.8 & 0.9 \\
 & UCCH & 22.4 & \multicolumn{1}{c|}{23.5} & 22.4 & \multicolumn{1}{c|}{23.5} & 22.4 & \multicolumn{1}{c|}{23.5} & 22.4 & 23.5 & 5.5 & \multicolumn{1}{c|}{6.2} & 5.5 & \multicolumn{1}{c|}{6.2} & 5.5 & \multicolumn{1}{c|}{6.2} & 5.5 & 6.2 \\
 & GNN4CMR & 27.0 & \multicolumn{1}{c|}{26.0} & 21.0 & \multicolumn{1}{c|}{19.9} & 15.7 & \multicolumn{1}{c|}{13.8} & 14.2 & 11.4 & {\underline{36.2}} & \multicolumn{1}{c|}{{\underline{38.2}}} & {\underline{28.2}} & \multicolumn{1}{c|}{{\underline{30.5}}} & {\underline{21.0}} & \multicolumn{1}{c|}{{\underline{23.3}}} & {\underline{18.7}} & {\underline{21.5}} \\
 & RONO & 20.3 & \multicolumn{1}{c|}{19.3} & 18.9 & \multicolumn{1}{c|}{18.0} & 17.9 & \multicolumn{1}{c|}{16.8} & 17.7 & 16.5 & 17.1 & \multicolumn{1}{c|}{15.7} & 17.6 & \multicolumn{1}{c|}{15.8} & 17.2 & \multicolumn{1}{c|}{16.2} & 14.4 & 14.0 \\
 & CIRH & 11.7 & \multicolumn{1}{c|}{11.4} & 11.7 & \multicolumn{1}{c|}{11.4} & 11.7 & \multicolumn{1}{c|}{11.4} & 11.7 & 11.4 & 6.7 & \multicolumn{1}{c|}{6.6} & 6.7 & \multicolumn{1}{c|}{6.6} & 6.7 & \multicolumn{1}{c|}{6.6} & 6.7 & 6.6 \\
 & NCH & 1.7 & \multicolumn{1}{c|}{1.9} & 2.1 & \multicolumn{1}{c|}{2.1} & 1.9 & \multicolumn{1}{c|}{1.8} & 1.4 & 1.6 & 0.9 & \multicolumn{1}{c|}{1.2} & 0.8 & \multicolumn{1}{c|}{1.0} & 0.8 & \multicolumn{1}{c|}{1.0} & 0.8 & 0.9 \\
 & NRCH & 21.6 & \multicolumn{1}{c|}{22.0} & 20.9 & \multicolumn{1}{c|}{21.4} & 20.8 & \multicolumn{1}{c|}{21.4} & 21.3 & 21.4 & 9.8 & \multicolumn{1}{c|}{11.1} & 9.2 & \multicolumn{1}{c|}{10.6} & 9.2 & \multicolumn{1}{c|}{10.7} & 9.1 & 10.5 \\
 & CICH & 4.0 & \multicolumn{1}{c|}{3.9} & 4.0 & \multicolumn{1}{c|}{4.0} & 4.0 & \multicolumn{1}{c|}{4.0} & 4.0 & 4.0 & 1.4 & \multicolumn{1}{c|}{2.0} & 1.2 & \multicolumn{1}{c|}{1.7} & 0.9 & \multicolumn{1}{c|}{0.8} & 0.7 & 0.7 \\
 & RSHNL & {\underline{29.8}} & \multicolumn{1}{c|}{{\underline{29.9}}} & {\underline{27.6}} & \multicolumn{1}{c|}{{\underline{27.7}}} & {\underline{26.8}} & \multicolumn{1}{c|}{{\underline{26.5}}} & {\underline{25.3}} & {\underline{25.0}} & 21.8 & \multicolumn{1}{c|}{23.5} & 17.3 & \multicolumn{1}{c|}{18.7} & 13.6 & \multicolumn{1}{c|}{14.9} & 11.1 & 12.0 \\
 & DHRL & 14.2 & \multicolumn{1}{c|}{13.5} & 10.2 & \multicolumn{1}{c|}{10.7} & 4.3 & \multicolumn{1}{c|}{4.9} & 2.3 & 2.3 & 8.1 & \multicolumn{1}{c|}{9.3} & 4.6 & \multicolumn{1}{c|}{5.2} & 2.4 & \multicolumn{1}{c|}{2.9} & 1.1 & 1.2 \\
 & EGICH & 25.5 & \multicolumn{1}{c|}{25.2} & 18.8 & \multicolumn{1}{c|}{18.3} & 15.4 & \multicolumn{1}{c|}{15.0} & 12.4 & 11.9 & 28.1 & \multicolumn{1}{c|}{30.3} & 22.4 & \multicolumn{1}{c|}{24.8} & 19.2 & \multicolumn{1}{c|}{21.5} & 15.0 & 16.8 \\
 & SPIRN & \textbf{43.5} & \multicolumn{1}{c|}{\textbf{44.9}} & \textbf{39.4} & \multicolumn{1}{c|}{\textbf{40.2}} & \textbf{35.5} & \multicolumn{1}{c|}{\textbf{36.4}} & \textbf{28.2} & \textbf{28.8} & \textbf{49.8} & \multicolumn{1}{c|}{\textbf{48.9}} & \textbf{49.6} & \multicolumn{1}{c|}{\textbf{49.0}} & \textbf{49.2} & \multicolumn{1}{c|}{\textbf{48.5}} & \textbf{37.1} & \textbf{36.0} \\ \hline
\end{tabular}
}
\caption{Comparison of mAP (\%) on INRIA-Websearch and XMediaNet under different completeness and noise settings.}
\label{tab:inria_xmedianet_comparison_results}
\end{table*}

\subsection{Quality-aware Robust Learning}
According to the early learning phenomenon \cite{RSHNL}, deep networks tend to fit simple, clean patterns before memorizing corrupted labels. During early training, clean and noisy samples occupy distinct regions in the continuous semantic space. Motivated by this observation, we design a Quality-aware Robust Learning (QRL) module to assess sample quality and dynamically handle noisy data.

To effectively distinguish clean samples from noisy ones, we utilize the cross-entropy loss relative to the semantic priors as a quality metric. Let $\ell_i$ denote the semantic alignment loss for the $i$-th sample. To avoid relying on a manually predefined hard threshold, we propose a dynamic mechanism anchored by the cleanest samples in the current mini-batch $\mathcal{B}$. At each training step $t$, the dynamic threshold $\gamma_t$ is adaptively determined by the minimum loss within the batch:
\begin{equation}
    \gamma_t = \eta \cdot \min_{j \in \mathcal{B}} (\ell_j) + \epsilon,
    \label{eq:4}
\end{equation}
where $\eta$ is a predefined hyperparameter controlling the tolerance for noise, and $\epsilon$ is a small constant to prevent numerical instability. Based on this dynamic threshold, we assign a continuous quality-aware weight $w_i$ to each sample:
\begin{equation}
    w_i = \max \left( 0, 1 - \frac{\ell_i}{\gamma_t} \right).
    \label{eq:5}
\end{equation}
This weighting mechanism intuitively isolates the noise, and high-quality samples (i.e., $\ell_i \ll \gamma_t$) are assigned larger weights and pulled closer to their assigned semantic priors. This is achieved by minimizing the following weighted semantic alignment loss:
\begin{equation}
    \mathcal{L}_{s} =
    \frac{\sum_{i \in \mathcal{B}} w_i \ell_i}
    {\sum_{i \in \mathcal{B}} w_i}.
    \label{eq:6}
\end{equation}
Conversely, samples with losses exceeding $\gamma_t$ are treated as noisy and assigned zero weight. Instead of discarding them, we employ a reverse learning loss to leverage their potentially useful structural information. For each detected noisy sample with observed label $\tilde{c}$, we randomly select a class $\hat{c} \neq \tilde{c}$ and penalize the model for predicting that class:
\begin{equation}
    \mathcal{L}_{r} =
    - \sum_{i \in \mathcal{B}} \mathbb{I}(w_i = 0)
    \sum_{* \in \{v,u\}}
    \log \left(1 - p(\hat{c} \mid x_i^*)\right),
    \label{eq:7}
\end{equation}
where $\mathbb{I}(\cdot)$ is the indicator function. 
Furthermore, to exploit cross-modal correspondences in the complete dataset $\mathcal{F}$ and reduce modality heterogeneity, we introduce a cross-modal contrastive objective. For an image-text pair $(\mathbf{z}_i^v, \mathbf{z}_i^u)$, we pull matched representations closer and push mismatched pairs apart using a symmetric InfoNCE loss. Let $\mathcal{P} = \{(v,u),(u,v)\}$ denote the two directional modality pairs:
\begin{equation}
\mathcal{L}_{c} =
- \frac{1}{2|\mathcal{B}|}
\sum_i \sum_{(m,n)\in\mathcal{P}}
\log
\frac{\exp\!\left((\mathbf{z}_i^m)^\top \mathbf{z}_i^n / \tau_c\right)}
{\sum_j \exp\!\left((\mathbf{z}_i^m)^\top \mathbf{z}_j^n / \tau_c\right)},
\label{eq:8}
\end{equation}
where $i,j \in \mathcal{B}$ and $\tau_c$ is the temperature hyperparameter. The overall objective function for this stage is:
\begin{equation}
    \mathcal{L}_{\mathrm{noisy}} =
    \mathcal{L}_{s} + \alpha \mathcal{L}_{r} + \beta \mathcal{L}_{c},
    \label{eq:9}
\end{equation}
where $\alpha$ and $\beta$ are trade-off hyperparameters that balance the contributions of the reverse learning and the contrastive learning, respectively.

\subsection{Prior-guided Global Label Calibration}
After quality-aware robust learning, the model establishes a robust continuous semantic space that substantially mitigates the impact of noisy annotations. However, directly using the original corrupted labels $\tilde{\mathbf{l}}_i$ to supervise subsequent missing-modality imputation may introduce semantic ambiguity and propagate errors. To provide reliable guidance for unobserved modalities, we design a Prior-guided Global Label Calibration (PGLC) module.

The fixed orthogonal anchors $\mathbf{W}$ provide stable semantic references for re-evaluating the category of each training instance. Specifically, we obtain robust continuous representations for all instances using the optimized encoders. For each instance in the complete multimodal set $\mathcal{F}$, we aggregate and $L_2$-normalize its visual and textual embeddings to obtain a unified representation $\hat{\mathbf{z}}_i$:
\begin{equation}
    \hat{\mathbf{z}}_i =
    \frac{\mathbf{z}_i^v + \mathbf{z}_i^u}
    {\|\mathbf{z}_i^v + \mathbf{z}_i^u\|_2},
    \; \forall i \in \mathcal{F}.
    \label{eq:10}
\end{equation}
For the incomplete unimodal sets $\mathcal{V}$ and $\mathcal{U}$, we directly use the embedding $\mathbf{z}_i^v$ or $\mathbf{z}_i^u$ as $\hat{\mathbf{z}}_i$.
Subsequently, we compute the cosine similarities between the unified representation $\hat{\mathbf{z}}_i$ and the semantic priors in $\mathbf{W}$. The rectified pseudo-label $\hat{c}_i$ is assigned to the class with the highest similarity:

\begin{equation}
    \hat{c}_i =
    \mathop{\arg\max}_{c \in \{1,2,\dots,C\}}
    \left( \hat{\mathbf{z}}_i^\top \mathbf{w}_c \right).
    \label{eq:11}
\end{equation}
By executing this global rectification, we obtain refined one-hot labels $\hat{\mathbf{l}}_i \in \{0,1\}^C$ for the entire dataset $\mathcal{O}$.

\subsection{Confidence-driven Selective Imputation}
Bridging the semantic gap is central to cross-modal retrieval. However, the lack of explicit correspondences in unpaired data severely hinders alignment. Although missing modalities can be imputed from available data, label noise may yield embeddings of varying quality and semantic deviations. To address this issue, we design a Confidence-driven Selective Imputation (CSI) module that progressively recovers missing modalities while filtering unreliable imputations. 

First, we construct a reliable cross-modal memory bank from the refined representations of the complete multimodal set $\mathcal{F}$. For an image-only instance $i \in \mathcal{V}$ with embedding $\mathbf{z}_i^v$, we retrieve its $K$ nearest neighbors $\mathcal{N}_K(i)$ from the visual memory bank based on cosine similarity. The missing textual embedding $\hat{\mathbf{z}}_i^u$ is then imputed by a similarity-weighted aggregation of the corresponding textual embeddings:
\begin{equation}
    a_{i,j} =
    \frac{\exp((\mathbf{z}_i^v)^\top \mathbf{z}_j^v / \tau)}
    {\sum_{k \in \mathcal{N}_K(i)} \exp((\mathbf{z}_i^v)^\top \mathbf{z}_k^v / \tau)},
    \label{eq:12}
\end{equation}
\begin{equation}
    \hat{\mathbf{z}}_i^u =
    \frac{\sum_{j \in \mathcal{N}_K(i)} a_{i,j}\mathbf{z}_j^u}
    {\left\|\sum_{j \in \mathcal{N}_K(i)} a_{i,j}\mathbf{z}_j^u\right\|_2},
    \label{eq:13}
\end{equation}
where $\tau$ is a temperature parameter. 
Subsequently, we use the semantic priors to assess imputation quality. We compute the confidence $t_i$ that the imputed embedding $\hat{\mathbf{z}}_i^u$ belongs to its rectified label $\hat{c}_i$ using a sigmoid function $\sigma(\cdot)$, and define its uncertainty score as:
\begin{equation}
    s_i = 1 - t_i =
    1 - \sigma\!\left(
    (\hat{\mathbf{z}}_i^u)^\top \mathbf{w}_{\hat{c}_i} / \tau
    \right).
    \label{eq:14}
\end{equation}
To prevent low-quality imputations from interfering with training, we use a dynamically increasing threshold $\gamma_m$ to control their introduction. A continuous confidence-aware weight $\hat{w}_i$ is then assigned to each imputed embedding:
\begin{equation}
    \hat{w}_i =
    \max\left(0, 1 - \frac{s_i}{\gamma_m}\right).
    \label{eq:15}
\end{equation}
Finally, only reliable imputations ($\hat{w}_i > 0$) are incorporated into training. The model is optimized by pulling both observed and imputed embeddings towards their semantic anchors while aligning cross-modal pairs:
\begin{equation}
\mathcal{L}_{\mathrm{miss}} =
\frac{
\sum_{i \in \mathcal{V}} \hat{w}_i \, \phi(\mathbf{z}_i^v, \hat{\mathbf{z}}_i^u, \hat{c}_i)
+
\sum_{j \in \mathcal{U}} \hat{w}_j \, \phi(\hat{\mathbf{z}}_j^v, \mathbf{z}_j^u, \hat{c}_j)
}{
\sum_{k \in \mathcal{V} \cup \mathcal{U}} \hat{w}_k
}.
\label{eq:16}
\end{equation}
For an imputed pair, we define
$\phi(\mathbf{a}, \mathbf{b}, c)
= \ell(\mathbf{a}, c) + \ell(\mathbf{b}, c)
+ \beta \ell_c(\mathbf{a}, \mathbf{b})$,
where $\ell(\cdot, \cdot)$ and $\ell_c(\cdot, \cdot)$ denote the semantic alignment loss and pairwise contrastive term. 

\subsection{Optimization}
Guided by semantic consistency, the training process comprises three core stages: Warm-up, Quality-aware Robust Learning, and Confidence-driven Selective Imputation. At epoch $e$, the total objective $\mathcal{L}_{\mathrm{total}}$ is defined as:
\begin{equation}
    \mathcal{L}_{\mathrm{total}} =
    \begin{cases}
        \mathcal{L}_{\mathrm{spg}}, & \text{if } e \le E_{\mathrm{warm}}, \\
        \mathcal{L}_{\mathrm{noisy}}, & \text{if } E_{\mathrm{warm}} < e \le E_{\mathrm{noisy}}, \\
        \mathcal{L}_{\mathrm{miss}}, & \text{if } E_{\mathrm{noisy}} < e \le E_{\mathrm{max}}.
    \end{cases}
    \label{eq:optimization}
\end{equation}
where $E_{\mathrm{warm}}$, $E_{\mathrm{noisy}}$, and $E_{\mathrm{max}}$ denote the epoch thresholds for the Warm-up, Quality-aware Robust Learning, and Confidence-driven Selective Imputation stages, respectively.

\section{Experiments}
\subsection{Dataset}
In our experiments, we evaluate the proposed SPIRN framework on four widely used benchmark datasets, i.e., Wikipedia \cite{rasiwasia2010new}, XMedia \cite{peng2015semi}, INRIA-Websearch \cite{krapac2010improving}, and XMediaNet \cite{peng2017overview}. 
These datasets all contain image-text pairs and corresponding labels.

\begin{figure}[t]
    \centering
    \makebox[\columnwidth][c]{%
        \includegraphics[width=0.38\columnwidth]{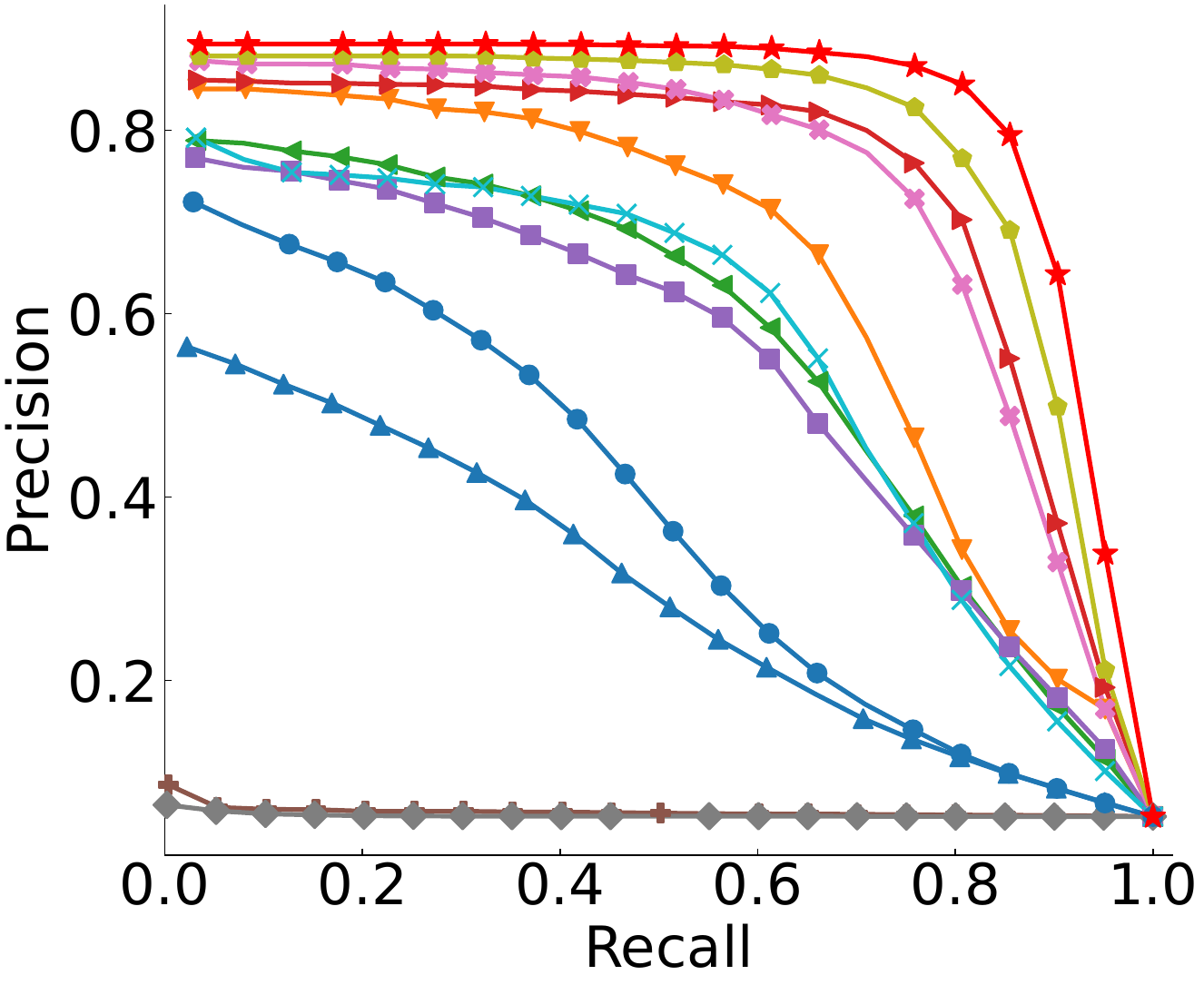}\hspace{0.02\columnwidth}
        \includegraphics[width=0.49\columnwidth]{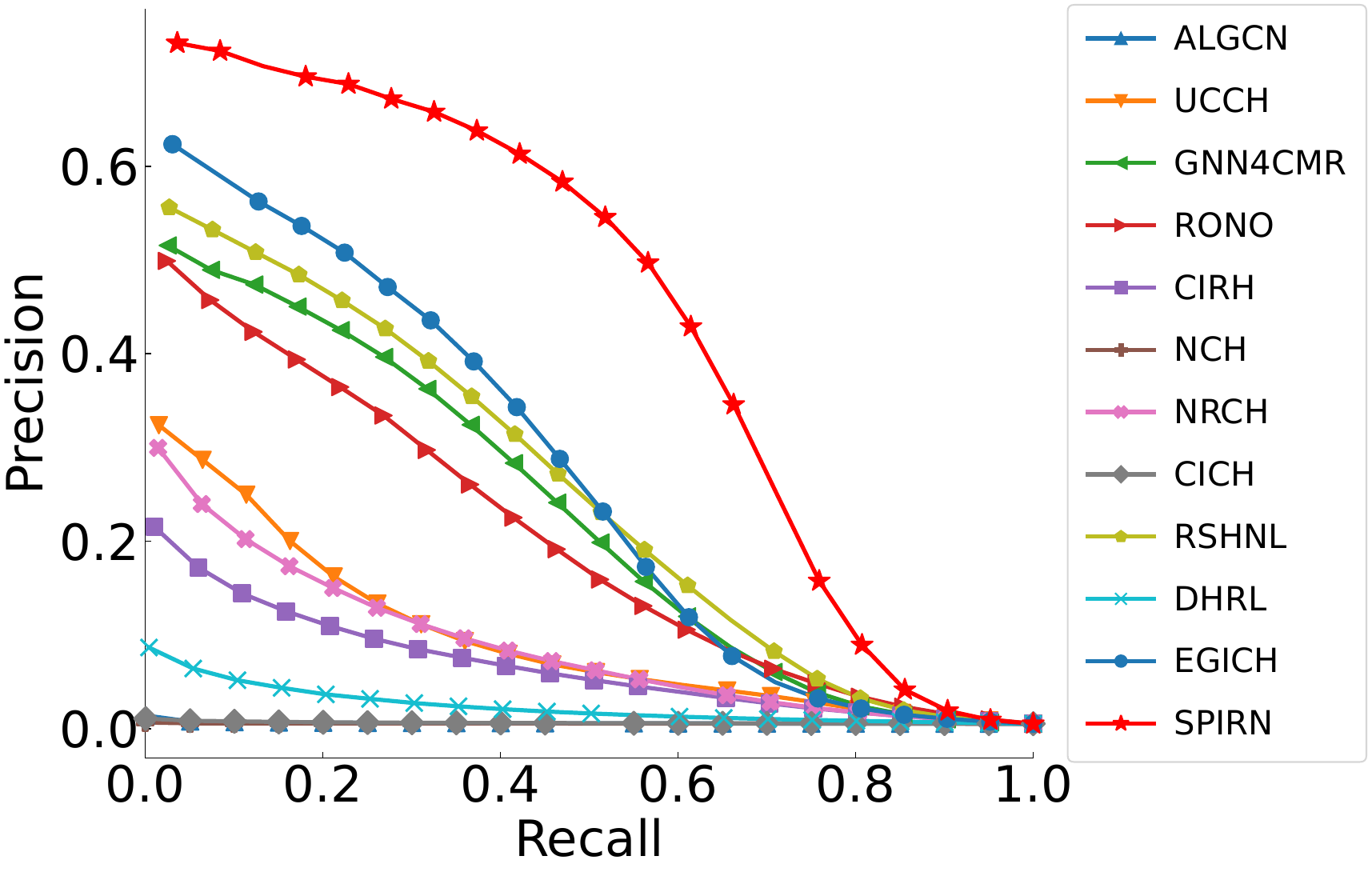}
    }

    \makebox[\columnwidth][c]{%
        \footnotesize (a) XMedia (I2T)\hspace{0.18\columnwidth}
        \footnotesize (b) XMediaNet (I2T)
    }

    \vspace{2pt}

    \makebox[\columnwidth][c]{%
        \includegraphics[width=0.38\columnwidth]{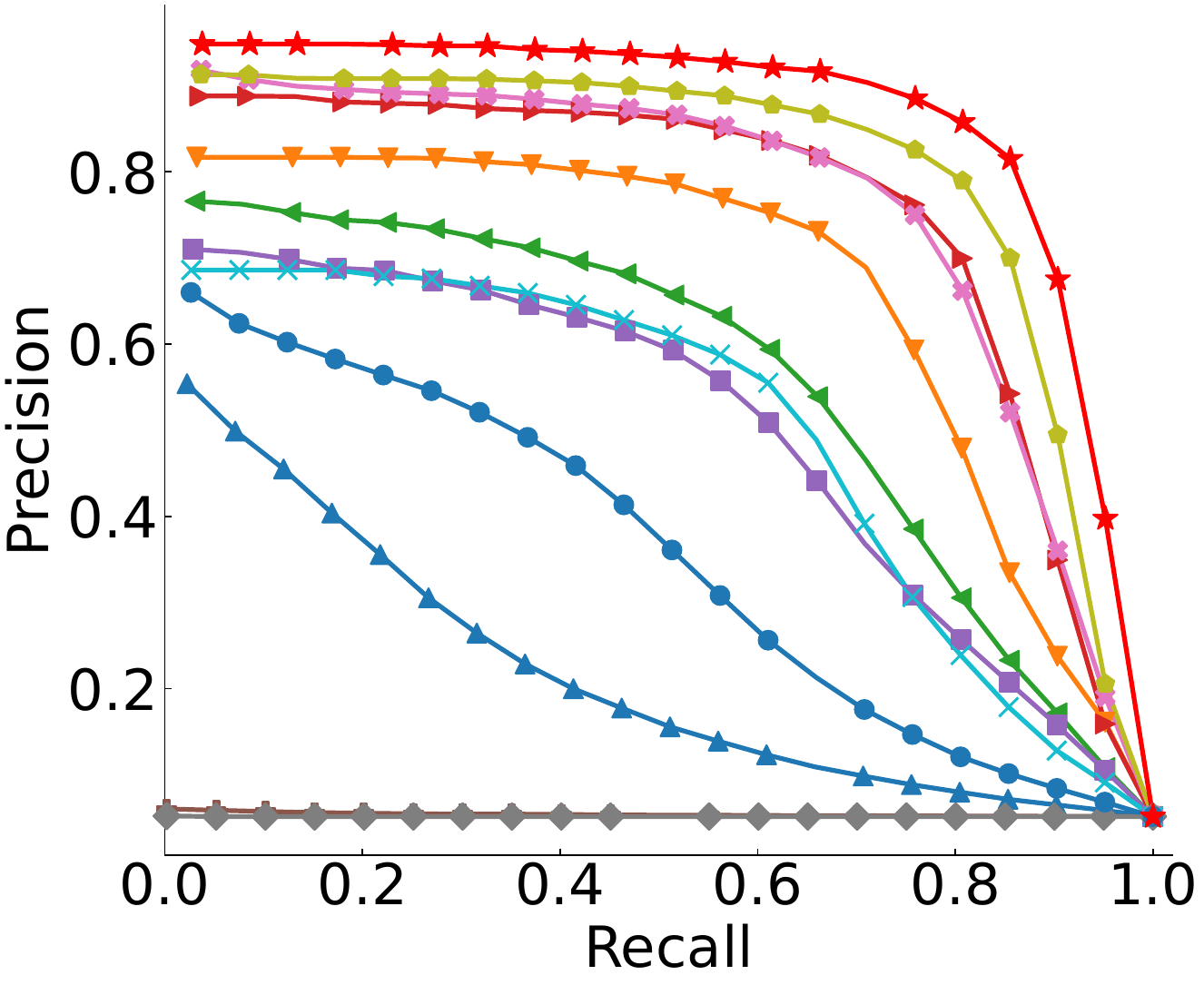}\hspace{0.02\columnwidth}
        \includegraphics[width=0.49\columnwidth]{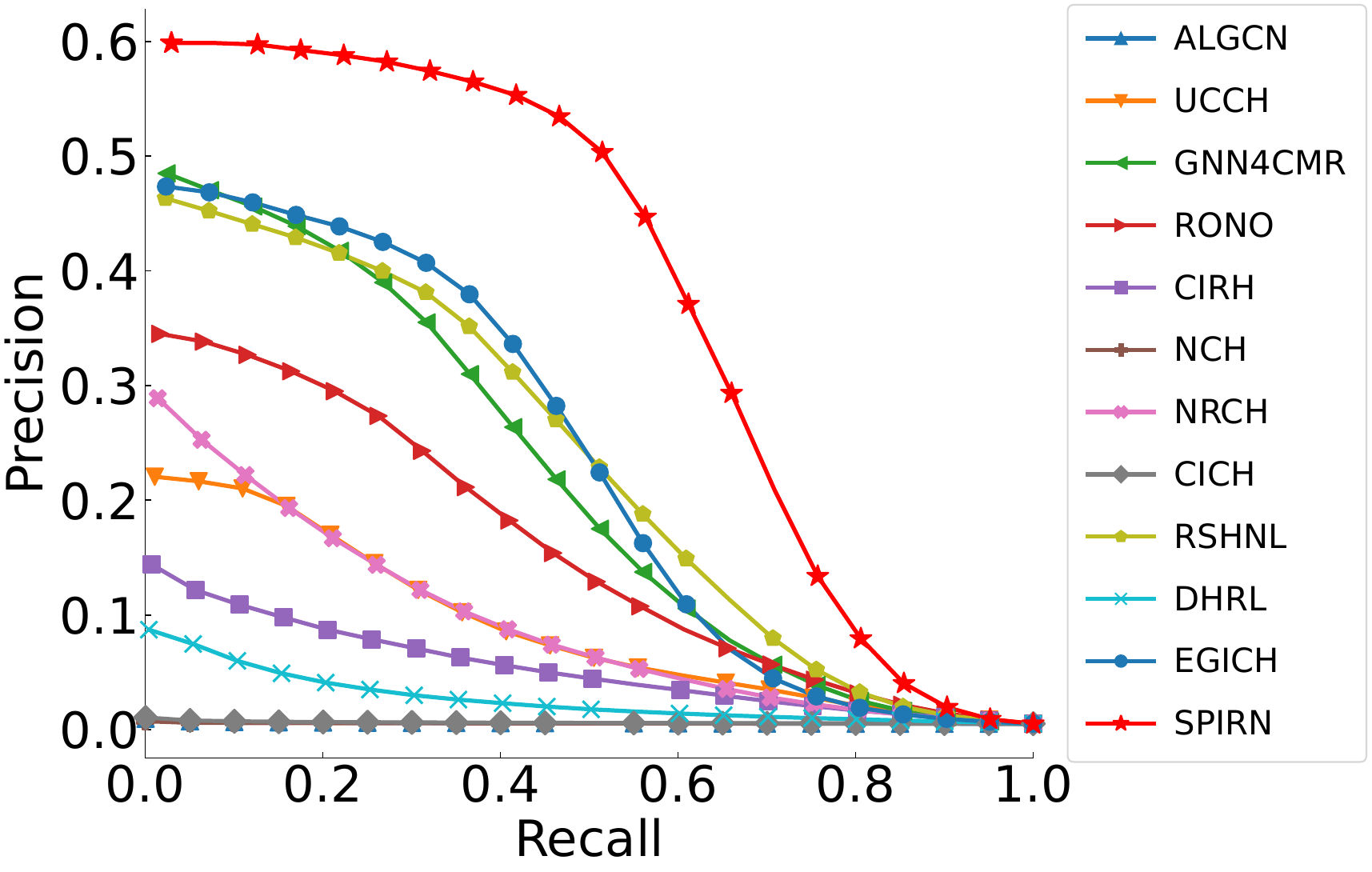}
    }

    \makebox[\columnwidth][c]{%
        \footnotesize (c) XMedia (T2I)\hspace{0.18\columnwidth}
        \footnotesize (d) XMediaNet (T2I)
    }

    \caption{Precision-recall curves on XMedia and XMediaNet under the medium missing setting with a noise ratio of 0.6.}
    \label{fig:pr}
\end{figure}

\subsection{Experimental Settings}
\textbf{Incomplete Modality Scenarios}: Following previous ICMR studies, we consider three settings of increasing severity, termed easy, medium, and hard. The ratios of complete, image-only, and text-only training samples are (80\%, 10\%, 10\%), (50\%, 25\%, 25\%), and (20\%, 40\%, 40\%), respectively.
\\
\textbf{Noisy Label Scenarios}: To evaluate robustness to label noise, we inject synthetic noise only into the training set and retain clean validation and test sets. Under symmetric random-label noise, we select 20\%, 40\%, 60\%, or 80\% of the training samples and uniformly resample their labels.
\\
\textbf{Implementation Details}: SPIRN is implemented in PyTorch and trained on a single GeForce RTX 4090 GPU. We use mean average precision (mAP)~\cite{cui2019scalable,li2019relaxed} as the primary metric and precision-recall (PR) curves for complementary evaluation.

\subsection{Comparison Methods}
To verify the superiority of SPIRN, we compare it with 11 SOTA baselines, including two unsupervised methods (i.e., UCCH \cite{UCCH} and CIRH \cite{CIRH}) and nine supervised methods (i.e., ALGCN \cite{ALGCN}, GNN4CMR \cite{GNN4CMR}, RONO \cite{feng2023rono}, NCH \cite{NCH}, NRCH \cite{NRCH}, CICH \cite{luo2024contrastive}, RSHNL \cite{RSHNL}, DHRL \cite{DHRL}, and EGICH \cite{EGICH}).

\subsection{Comparisons with State-of-the-Art Methods}
\textbf{Quantitative Comparisons:} To evaluate the effectiveness of SPIRN, we compare it with 11 methods on four datasets under missing-data settings and noise ratios. The results are reported in Tab.~\ref{tab:wiki_xmedia_comparison_results} and Tab.~\ref{tab:inria_xmedianet_comparison_results}.
We make the following observations: 1) As the noise ratio and missing rate increase, retrieval performance generally declines across most methods. 2) Although RONO, NRCH, RSHNL, and DHRL are robust to noisy labels, they are less effective at exploiting incomplete samples under modality missingness. 3) Although GNN4CMR, NCH, CICH, and EGICH are designed for incomplete multimodal data, their reliance on clean labels makes them vulnerable to severe label noise. 4) SPIRN achieves the best performance in nearly all settings by jointly selecting reliable samples and filtering low-quality imputations, with particularly clear gains in challenging scenarios.
\\
\textbf{Qualitative Comparisons:} To further evaluate the retrieval performance of SPIRN against 11 state-of-the-art methods, we present the Precision-Recall (PR) curves in Fig.~\ref{fig:pr}. SPIRN consistently maintains higher precision across all recall levels, further demonstrating its effectiveness.

\begin{table}[!t]
\centering
{\small
\begin{tabular}{c|ccc|ccc}
\hline
 & \multicolumn{3}{c|}{XMedia} & \multicolumn{3}{c}{XMediaNet} \\ \cline{2-7} 
\multirow{-2}{*}{Methods} & \multicolumn{1}{c|}{I2T} & \multicolumn{1}{c|}{T2I} & Mean & \multicolumn{1}{c|}{I2T} & \multicolumn{1}{c|}{T2I} & Mean \\ \hline
SPIRN-1 & \multicolumn{1}{c|}{83.97} & \multicolumn{1}{c|}{84.27} & 84.12 & \multicolumn{1}{c|}{15.64} & \multicolumn{1}{c|}{16.65} & 16.15 \\ \hline
SPIRN-2 & \multicolumn{1}{c|}{89.57} & \multicolumn{1}{c|}{89.81} & 89.69 & \multicolumn{1}{c|}{55.43} & \multicolumn{1}{c|}{55.68} & 55.56 \\ \hline
SPIRN-3 & \multicolumn{1}{c|}{82.65} & \multicolumn{1}{c|}{83.07} & 82.86 & \multicolumn{1}{c|}{55.18} & \multicolumn{1}{c|}{54.80} & 54.99 \\ \hline
SPIRN & \multicolumn{1}{c|}{\textbf{89.68}} & \multicolumn{1}{c|}{\textbf{90.00}} & \textbf{89.84} & \multicolumn{1}{c|}{ \textbf{55.79}} & \multicolumn{1}{c|}{ \textbf{55.82}} & \textbf{55.81} \\ \hline
\end{tabular}
}
\caption{The mAP (\%) results of the ablation study on the XMedia and XMediaNet datasets.}
\label{tab:ablation_results}
\end{table}

\begin{table}[!t]
\centering
{\small
\setlength{\tabcolsep}{4.5pt}
\begin{tabular}{c|ccc|ccc}
\hline
\multirow{2}{*}{Clean (\%)} & \multicolumn{3}{c|}{XMedia} & \multicolumn{3}{c}{XMediaNet} \\ \cline{2-7}
& \multicolumn{1}{c|}{Easy} & \multicolumn{1}{c|}{Medium} & Hard & \multicolumn{1}{c|}{Easy} & \multicolumn{1}{c|}{Medium} & Hard \\ \hline
80.0 & \multicolumn{1}{c|}{98.6} & \multicolumn{1}{c|}{96.9} & 94.0 & \multicolumn{1}{c|}{96.9} & \multicolumn{1}{c|}{91.3} & 80.6 \\ \hline
60.0 & \multicolumn{1}{c|}{98.3} & \multicolumn{1}{c|}{96.4} & 91.0 & \multicolumn{1}{c|}{96.9} & \multicolumn{1}{c|}{91.0} & 80.9 \\ \hline
40.0 & \multicolumn{1}{c|}{97.8} & \multicolumn{1}{c|}{95.9} & 91.3 & \multicolumn{1}{c|}{96.4} & \multicolumn{1}{c|}{90.3} & 80.5 \\ \hline
20.0 & \multicolumn{1}{c|}{96.8} & \multicolumn{1}{c|}{93.4} & 83.6 & \multicolumn{1}{c|}{90.7} & \multicolumn{1}{c|}{81.4} & 68.8 \\ \hline
\end{tabular}
}
\caption{The accuracy (\%) of global label calibration under different label noise and missing data settings.}
\label{tab:label_calibration}
\end{table}

\subsection{Ablation Study}
To assess the contribution of each component, we conduct an ablation study under the medium missing setting with a 0.6 label noise ratio on the XMedia and XMediaNet datasets. The results are reported in Tab.~\ref{tab:ablation_results}. We make the following observations: 1) \textbf{SPIRN-1}, which removes $\mathcal{L}_{s}$, leads to a substantial performance drop, indicating that semantic-prior guided alignment is essential for anchoring image and text embeddings in a shared semantic space. Without this supervisory signal, the model cannot reliably measure sample-prior consistency, making clean-noisy sample separation difficult. 2) \textbf{SPIRN-2}, which removes $\mathcal{L}_{r}$, also degrades performance, showing that reverse learning helps exploit noisy samples instead of simply discarding them, thereby preserving useful structural information. 3) \textbf{SPIRN-3}, which removes $\mathcal{L}_{c}$, further reduces performance, suggesting that semantic alignment alone is insufficient for fine-grained instance-level matching, while contrastive learning strengthens cross-modal correspondence. Overall, removing any component degrades performance, demonstrating the effectiveness of the proposed design.

\begin{figure}[!t]
\centering
\begin{tabular}{cc}
\includegraphics[width=0.42\columnwidth]{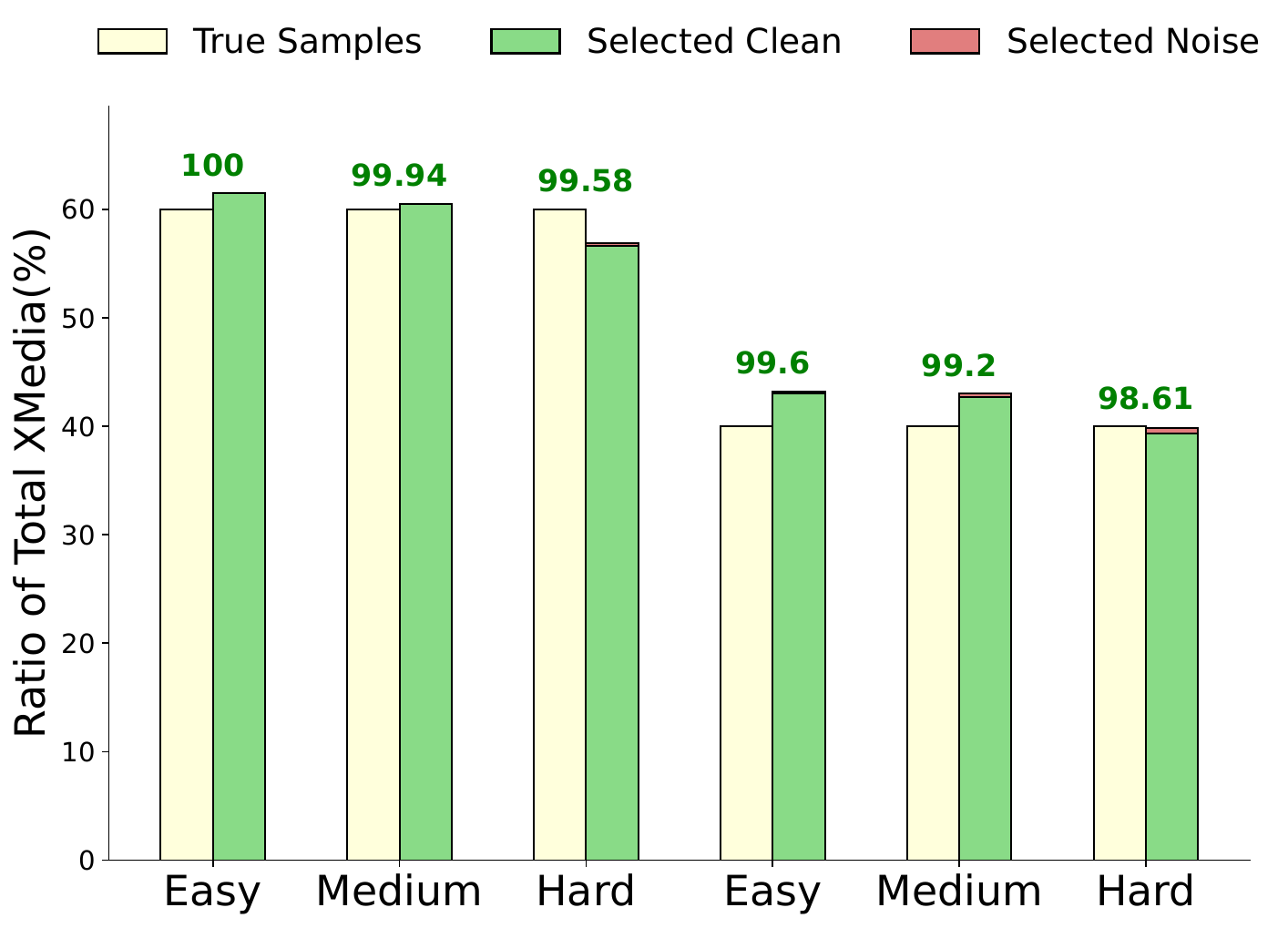} &
\includegraphics[width=0.42\columnwidth]{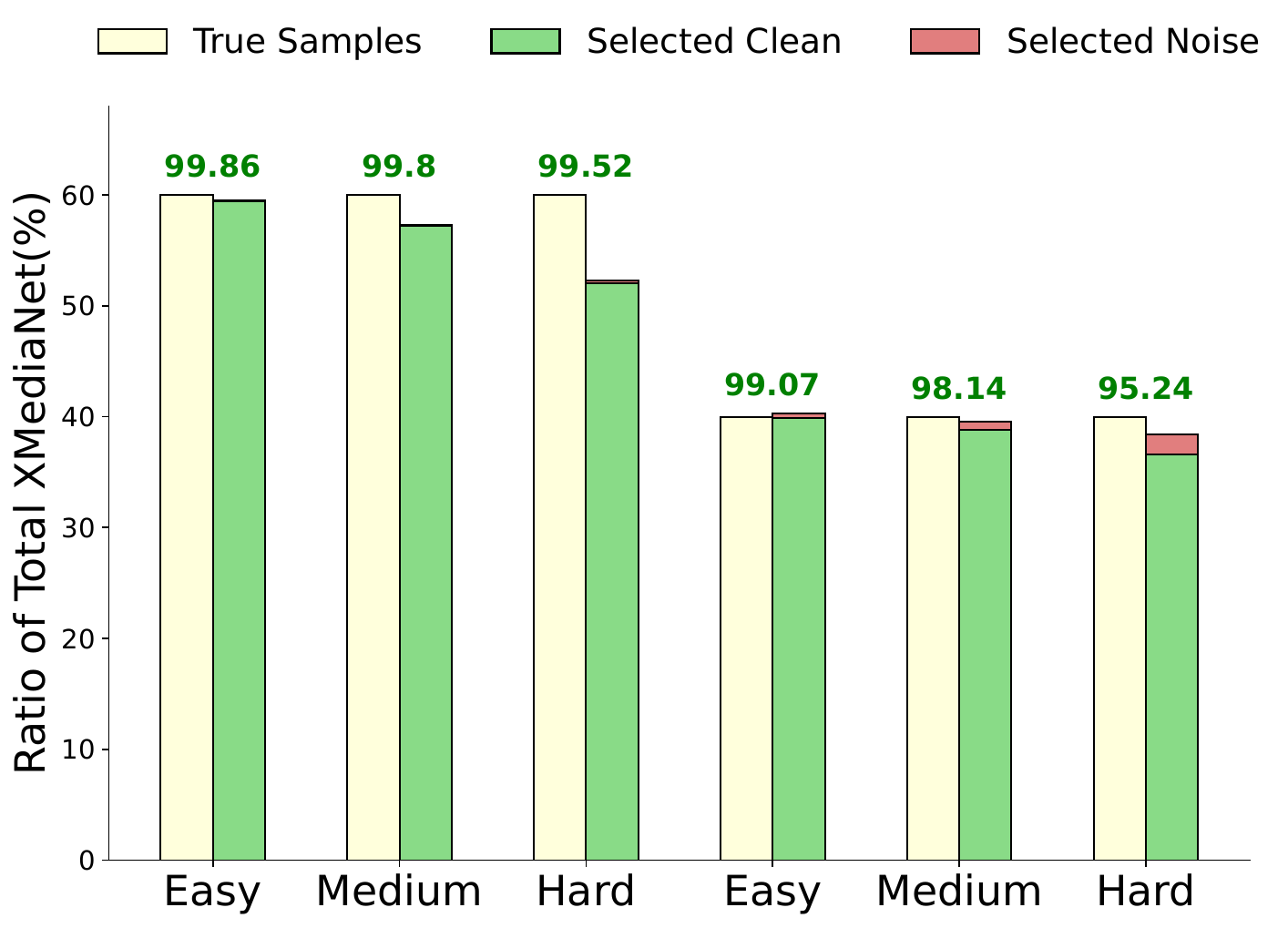} \\
\footnotesize (a) XMedia & \footnotesize (b) XMediaNet
\end{tabular}
\caption{The ratio (\%) of selected clean samples under different missing settings.}
\label{fig:noisy_select}
\end{figure}

\begin{figure}[!t]
    \centering
    \begin{tabular}{cc}
        \includegraphics[width=0.42\columnwidth]{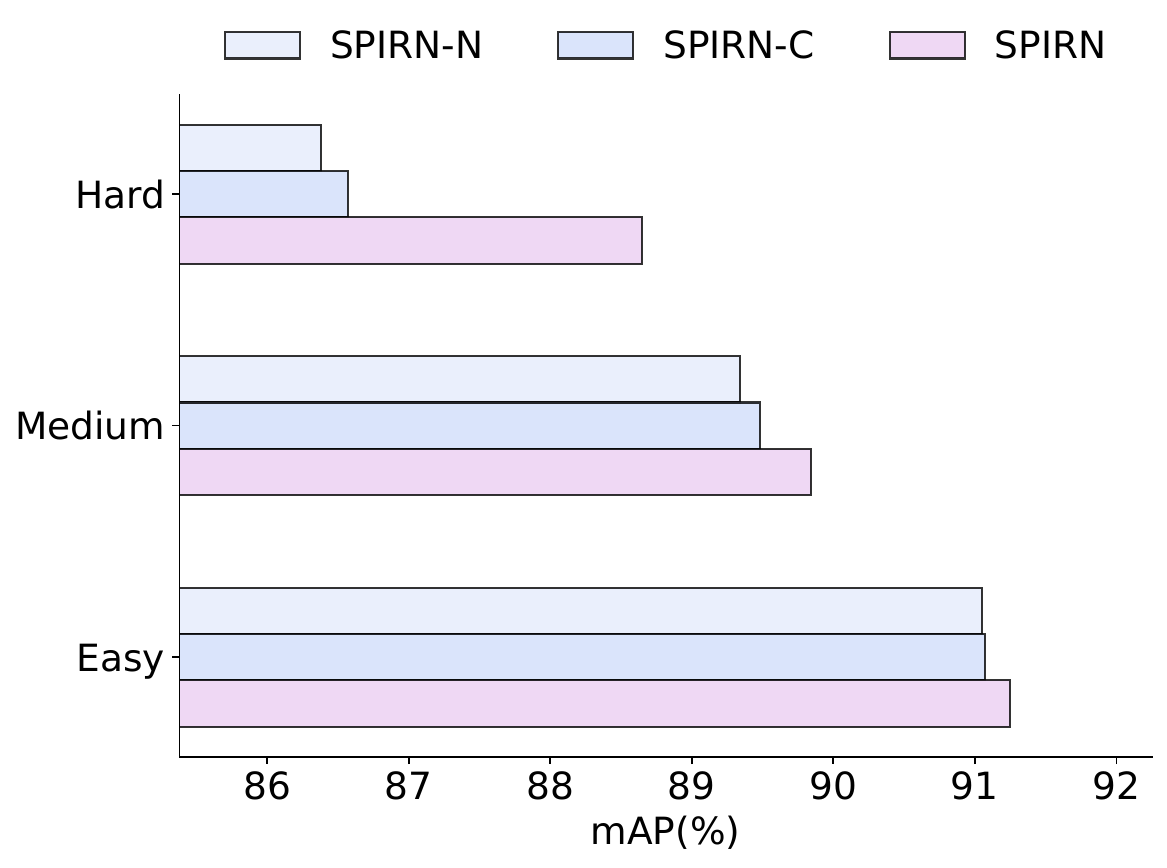} &
        \includegraphics[width=0.42\columnwidth]{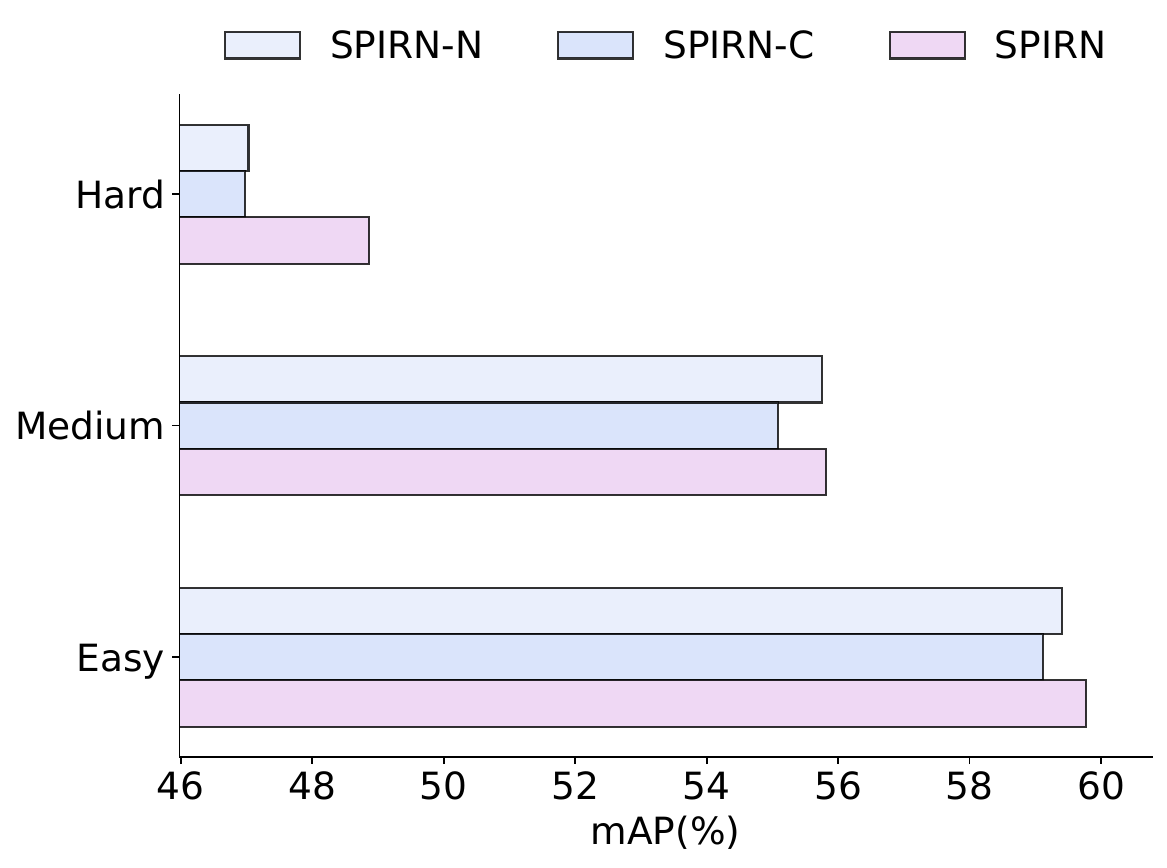} \\
        \footnotesize (a) XMedia & \footnotesize (b) XMediaNet
    \end{tabular}
    \caption{The mAP (\%) of three variants with the 0.6 noise ratio under different missing settings.}
    \label{fig:miss_select}
\end{figure}

\subsection{Component Analysis}
To investigate the contributions of three components, we conduct targeted experiments as follows: 1) \textbf{Quality-aware Robust Learning.} We evaluate the sample selection mechanism by visualizing the ratio of clean samples correctly identified by SPIRN under various label-noise and missing-data settings, as shown in Fig.~\ref{fig:noisy_select}. The results show that SPIRN accurately identifies clean samples across diverse settings, demonstrating robustness to noisy labels. 2) \textbf{Prior-guided Global Label Calibration.} We report label calibration accuracy under different settings in Tab.~\ref{tab:label_calibration}. The results indicate that SPIRN achieves effective global label calibration under diverse label-noise and missing-data conditions. 3) \textbf{Confidence-driven Selective Imputation.} We compare three variants: (a) \textit{SPIRN-N}, without embedding imputation; (b) \textit{SPIRN-C}, with full embedding imputation but without filtering low-quality samples; and (c) \textit{SPIRN}, the complete framework with confidence-driven selective imputation. As shown in Fig.~\ref{fig:miss_select}, selective imputation consistently improves performance, especially under severe missing-data and noisy-label settings. In contrast, indiscriminately using imputed embeddings may degrade performance.

\section{Conclusion}
In this paper, we propose SPIRN to address modality incompleteness and noisy labels in cross-modal retrieval. SPIRN consists of four key components: Semantic Priors Guidance (SPG), Quality-aware Robust Learning (QRL), Prior-guided Global Label Calibration (PGLC), and Confidence-driven Selective Imputation (CSI). By introducing fixed semantic priors, SPIRN establishes a stable shared semantic space for robust representation learning, noisy-label handling, and missing-modality imputation. To the best of our knowledge, SPIRN is the first framework to jointly address modality incompleteness and label noise for cross-modal retrieval. Extensive experiments on four benchmark datasets demonstrate the effectiveness and robustness of SPIRN, which consistently outperforms 11 state-of-the-art methods across various noise ratios and missing-data settings.

\bibliography{cite}

\clearpage
\section*{Supplementary Materials}
\section{Summary}
This supplementary material provides additional methodological details and comprehensive experimental evidence for SPIRN. It presents the symbol definitions, complete optimization procedure, dataset descriptions, implementation details, and baseline information. It further reports qualitative comparisons, random-seed analysis, severe-missingness results, asymmetric label-noise analysis, parameter and epoch sensitivity analyses, training curves, computational-cost analysis, semantic-prior construction ablations, and t-SNE visualizations.

\begin{algorithm}[!t]
   \caption{The optimization pipeline for SPIRN.}
   \label{alg:algorithm}
   {\raggedright \textbf{Input:} Total dataset $\mathcal{O} = \mathcal{F} \cup \mathcal{V} \cup \mathcal{U}$; noisy labels $\tilde{L} = \{\tilde{\mathbf{l}}_i\}_{i=1}^N$; epoch thresholds $E_{warm}, E_{noisy}, E_{max}$; batch size $n_{b}$; learning rate $lr$; and hyperparameters $\alpha, \beta, \eta, K$.\par}
\begin{algorithmic}[1]
   \STATE Randomly initialize the network parameters $\Theta^{v}$ and $\Theta^{u}$.
   \STATE Generate a set of fixed, orthogonal semantic anchors $\mathbf{W} = [\mathbf{w}_1, \mathbf{w}_2, \dots, \mathbf{w}_C]^\top$.
   \FOR{$Epoch = 1,2,\dots,$ {\bfseries to} $E_{max}$}
      \IF{$Epoch \le E_{warm}$}
         \FOR{each mini-batch in $\mathcal{F}$} 
            \STATE Compute features $\mathbf{v}_i, \mathbf{u}_i$ by $E^v(\cdot, \Theta^v)$ and $E^u(\cdot, \Theta^u)$.
            \STATE Compute the loss $\mathcal{L}_{\mathrm{spg}}$ according to Eq.3.
            \STATE Update $\Theta^v, \Theta^u$ by computing gradients: $\Theta^* = \Theta^* - lr \frac{\partial \mathcal{L}_{\mathrm{spg}} }{\partial \Theta^*}$.
         \ENDFOR
      \ELSIF{$E_{warm} < Epoch \le E_{noisy}$}
         \FOR{each mini-batch in $\mathcal{F} \cup \mathcal{V} \cup \mathcal{U}$}
            \STATE Compute the loss $\ell_i$ and update threshold $\gamma_t$ by Eq.4.
            \STATE Assign weights $w_i$ by Eq.5.
            \STATE Compute the loss $\mathcal{L}_{\mathrm{noisy}}$ according to Eq.9.
            \STATE Update $\Theta^v, \Theta^u$ by computing gradients: $\Theta^* = \Theta^* - lr \frac{\partial \mathcal{L}_{\mathrm{noisy}} }{\partial \Theta^*}$.
         \ENDFOR
         \IF{$Epoch == E_{noisy}$}
            \STATE Extract comprehensive representations $\mathbf{\hat{z}}_i$ for all instances in $\mathcal{O}$ by Eq.10.
            \STATE Obtain rectified labels $\hat{L} = \{\hat{\mathbf{l}}_i\}$ by Eq.11.
         \ENDIF
      \ELSE
         \STATE Progressively update the dynamic threshold $\gamma_m$.
         \FOR{each mini-batch in $\mathcal{F} \cup \mathcal{V} \cup \mathcal{U}$}
            \STATE Build a cross-modal memory bank by using representations from $\mathcal{F}$.
            \STATE Retrieve $K$ nearest neighbors $\mathcal{N}_K(i)$ and impute missing modalities $\hat{\mathbf{u}}_i$ or $\hat{\mathbf{v}}_j$ by Eq.13.
            \STATE Compute uncertainty score $s_i$ and confidence-aware weight $\hat{w}_i$ by Eq. 14 and Eq.15.
            \STATE Compute the loss $\mathcal{L}_{\mathrm{miss}}$ according to Eq.16.
            \STATE Update $\Theta^v, \Theta^u$ by computing gradients: $\Theta^* = \Theta^* - lr \frac{\partial \mathcal{L}_{\mathrm{miss}} }{\partial \Theta^*}$.
         \ENDFOR
      \ENDIF
   \ENDFOR
\end{algorithmic}
{\raggedright \textbf{Output:} The optimized parameters $\Theta^{v}$ and $\Theta^{u}$.\par}
\end{algorithm}

\subsection{Symbol Definitions}
Tab.~\ref{tab:hyperparameters} summarizes the key hyperparameters used in SPIRN for reproducibility. 
For dataset-dependent settings, the batch size $n_b$ is set to $128$ on Wikipedia and $256$ on the other datasets. 
The contrastive loss weight $\beta$ is set to $0.2$ on Wikipedia and XMediaNet, $1$ on XMedia, and $2$ on INRIA-Websearch. 
The QRL threshold factor $\eta$ is set to $2.2$ on Wikipedia and XMedia, and $1.8$ on INRIA-Websearch and XMediaNet. 
Other dataset-specific settings, including input dimensions, category numbers, missing/noise ratios, and data splits, are reported in the experimental setup.

\subsection{Algorithm}
\label{sec:algorithm}
To provide an intuitive overview of the proposed SPIRN framework, we summarize the optimization pipeline in Algorithm \ref{alg:algorithm}.

\subsection{Datasets}
\label{sec:datasets}
To comprehensively evaluate the effectiveness and scalability of SPIRN, we conduct experiments on four widely used cross-modal retrieval datasets with progressively increasing numbers of semantic categories: Wikipedia (10 categories), XMedia (20 categories), INRIA-Websearch (100 categories), and XMediaNet (200 categories). This setting enables us to assess the model under semantic spaces of different granularities and complexities. The following provides detailed descriptions of these datasets.

\textbf{\textit{Wikipedia}} comprises 2,866 image-text pairs categorized into 10 semantic classes. For feature representation, image modalities are encoded as 4,096-dimensional vectors via a pre-trained VGG-19 \cite{simonyan2014very} model, while text modalities are represented by 300-dimensional vectors extracted using a pre-trained Doc2vec \cite{lau2016empirical} model. Following previous works \cite{DRCL}, the dataset is randomly partitioned into training, validation, and test sets containing 2,173, 231, and 462 pairs, respectively.

\textbf{\textit{XMedia}} is a comprehensive benchmark encompassing five modalities across 20 semantic categories. In our work, we specifically focus on a subset of 5,000 image-text pairs. For feature representation, images are encoded into 4,096-dimensional vectors using a pre-trained VGG-19 model, while text is represented by 3,000-dimensional Bag-of-Words (BOW) embeddings. Following previous works \cite{DRCL}, the selected data is randomly partitioned into training, validation, and test sets with 4,000, 500, and 500 pairs, respectively.

\begin{table}[!t]
\centering
\setlength{\tabcolsep}{3.5pt}
\renewcommand{\arraystretch}{1.08}
\begin{tabular}{clcc}
\toprule
Symbol & Description & Unit & Value \\
\midrule
$D$                  & Embedding size       & dimension & $512$ \\
$n_b$                & Batch size           & samples   & $128/256$ \\
$E_{\mathrm{warm}}$  & Warm-up endpoint     & epoch     & $10$ \\
$E_{\mathrm{noisy}}$ & Robust-stage endpoint& epoch     & $40$ \\
$E_{\max}$           & Total epochs         & epoch     & $100$ \\
$\alpha$             & Reverse-loss weight  & scalar    & $1.0$ \\
$\beta$              & Contrastive weight   & scalar    & $0.2/1/2$ \\
$\eta$               & QRL threshold factor & scalar    & $1.8/2.2$ \\
$\gamma_m$           & CSI threshold        & scalar    & $0{\rightarrow}0.3$ \\
$K$                  & Nearest neighbors    & neighbors & $5$ \\
$\tau$             & temperature    & scalar    & $0.29$ \\
Seed                 & Random seed          & integer   & $2025$ \\
\bottomrule
\end{tabular}
\caption{Key hyperparameter settings of SPIRN.}
\label{tab:hyperparameters}
\end{table}

\textbf{\textit{INRIA-Websearch}} is a large-scale benchmark containing over 70,000 image-text pairs across 353 semantic categories. For our experiments, we utilize a subset of 14,698 pairs spanning 100 semantic classes. Image modalities are represented by 4,096-dimensional features extracted via a pre-trained AlexNet \cite{krizhevsky2012imagenet}, while text modalities are represented by 1,000-dimensional features obtained using LDA \cite{hu2021learning}. Following previous works \cite{DRCL}, the subset is randomly partitioned into training, validation, and test sets with 9,000, 1,332, and 4,366 pairs, respectively.

\textbf{\textit{XMediaNet}} is a large dataset containing five modalities. Following previous work~\cite{peng2017overview}, we select 200 categories to preserve the category hierarchy and use 40,000 image-text pairs for the experiments. For feature representation, images are encoded into 4,096-dimensional vectors via a pre-trained VGG-19 model, while text is represented by 300-dimensional Doc2vec embeddings. Following previous works \cite{DRCL}, we randomly divide the subset into training, validation, and test sets containing 32,000, 4,000, and 4,000 pairs, respectively.

\begin{table}[t]
\centering
\begin{tabular}{lccc}
\toprule
Method & Reference & Noisy & Missing \\
\midrule
ALGCN   & TMM'22    & No  & No  \\
UCCH    & TPAMI'22  & No  & No  \\
GNN4CMR & TPAMI'22  & No  & Yes \\
RONO    & CVPR'23   & Yes & No  \\
CIRH    & TKDE'23   & No  & No  \\
NCH     & TMM'23    & No  & Yes \\
NRCH    & ACM MM'24 & Yes & No  \\
CICH    & TKDE'24   & No  & Yes \\
RSHNL   & AAAI'25   & Yes & No  \\
DHRL    & TBD'25    & Yes & No  \\
EGICH   & TIP'26    & No  & Yes \\
\bottomrule
\end{tabular}
\caption{The methods of 11 state-of-the-art baselines.}
\label{tab:comparison_methods}
\end{table}

\begin{figure*}[!t]
\centering
{
\setlength{\tabcolsep}{2pt}
\begin{tabular}{cccc}
\includegraphics[width=0.215\textwidth]{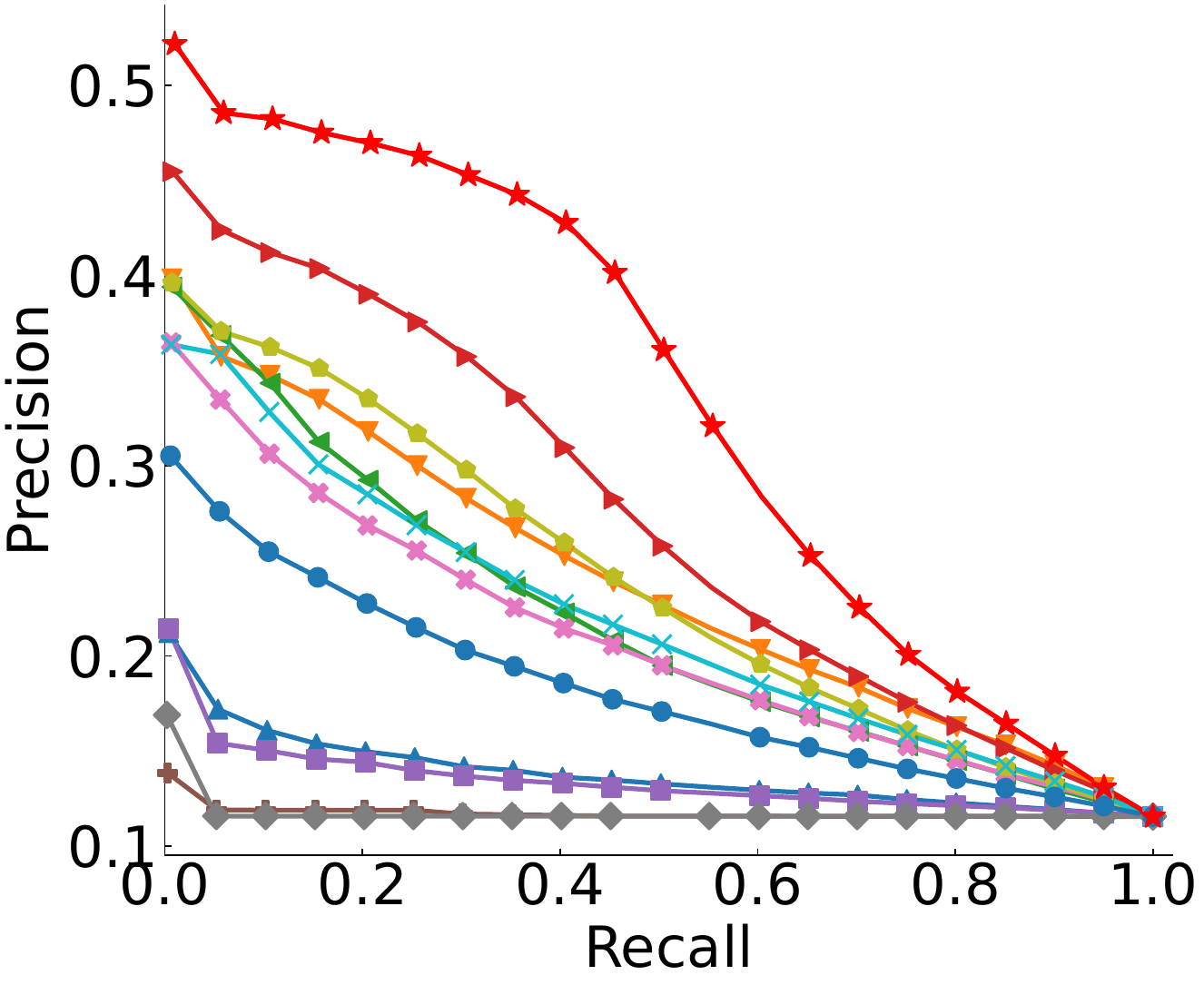} &
\includegraphics[width=0.215\textwidth]{Figures/xmedia_I2T_0.6PR.pdf} &
\includegraphics[width=0.215\textwidth]{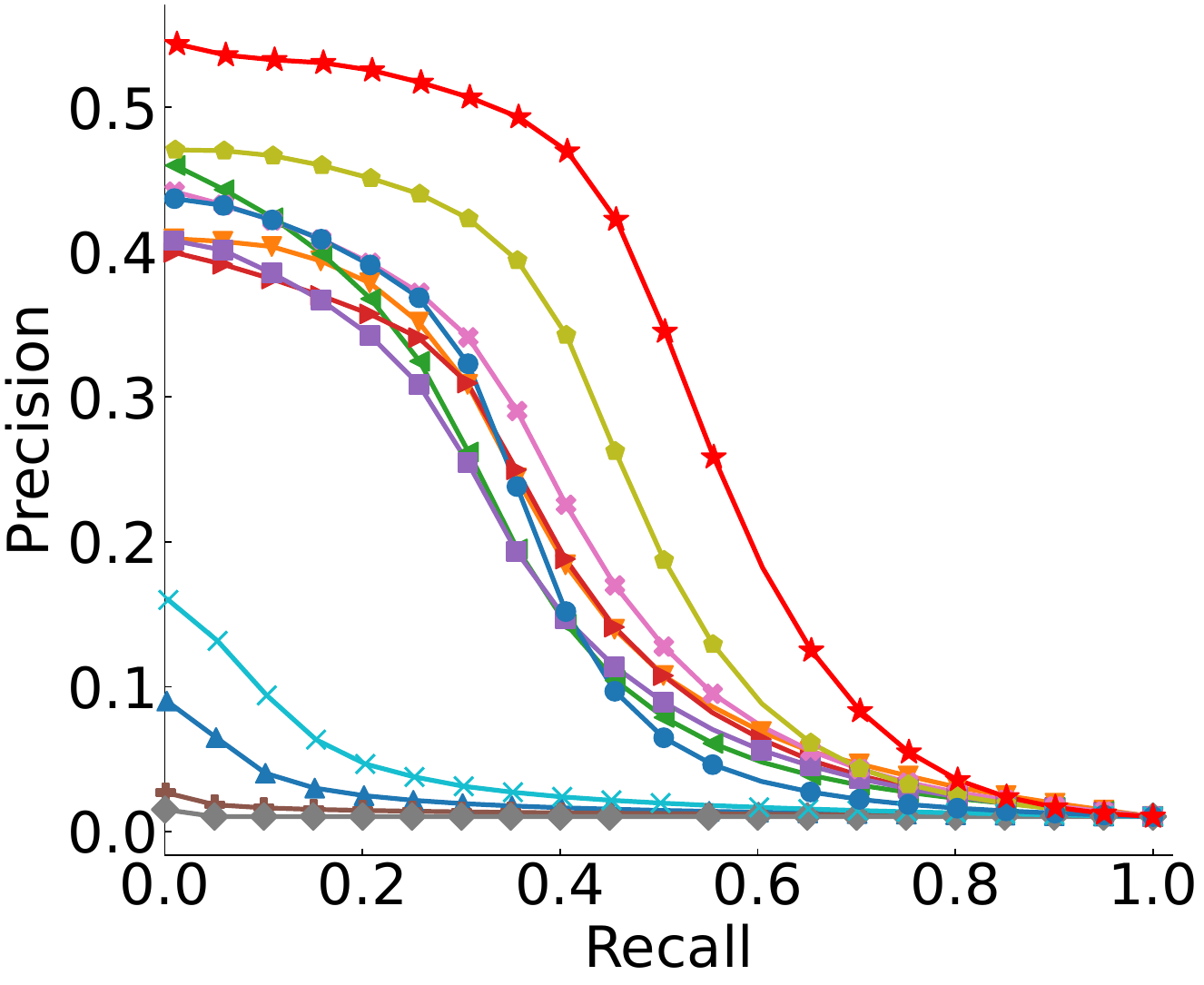} &
\includegraphics[width=0.28\textwidth]{Figures/xmedianet_I2T_0.6PR.pdf} \\
{\footnotesize (a) Wikipedia (I2T)} &
{\footnotesize (b) XMedia (I2T)} &
{\footnotesize (c) INRIA-Websearch (I2T)} &
{\footnotesize (d) XMediaNet (I2T)} \\
\includegraphics[width=0.215\textwidth]{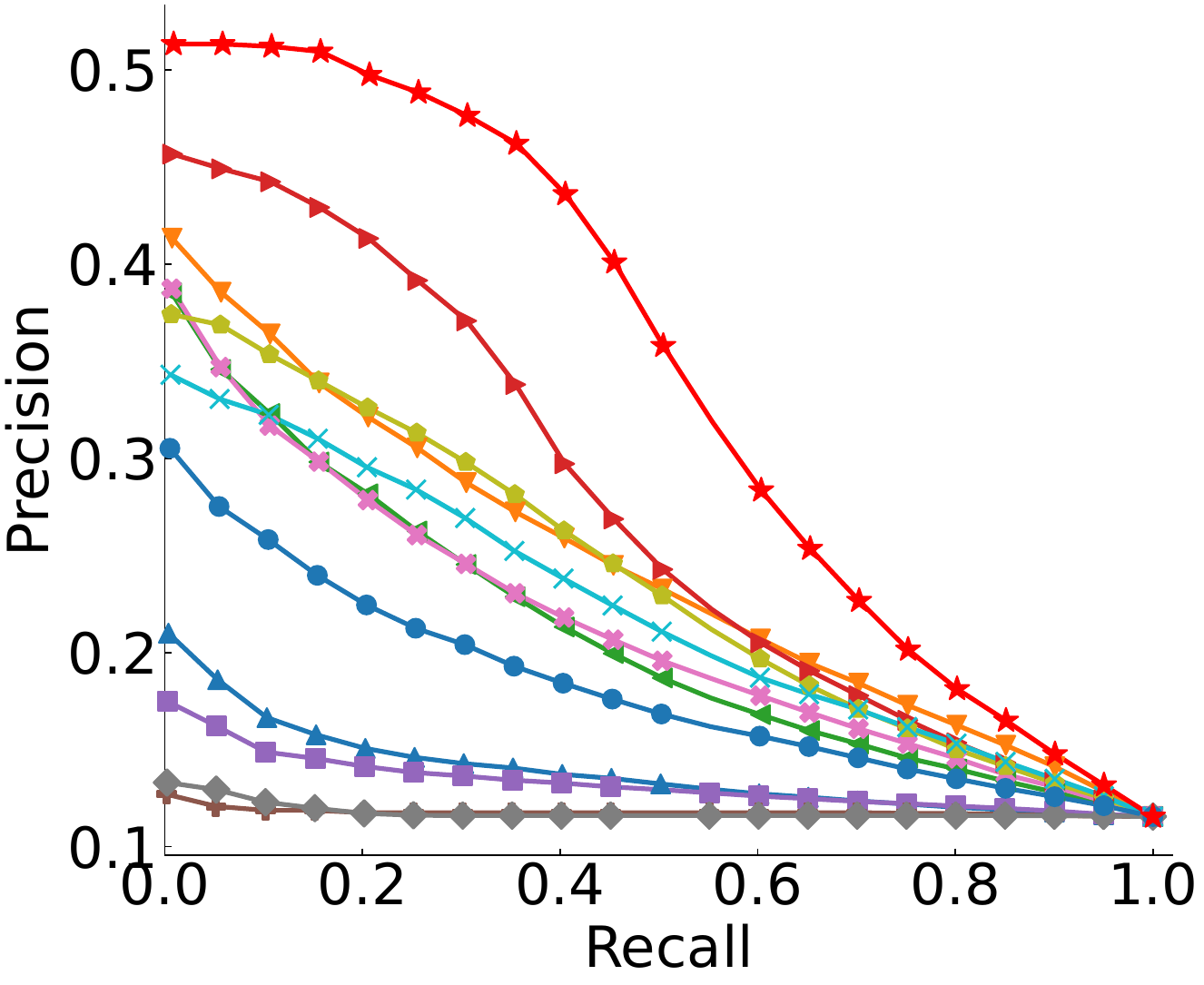} &
\includegraphics[width=0.215\textwidth]{Figures/xmedia_T2I_0.6PR.pdf} &
\includegraphics[width=0.215\textwidth]{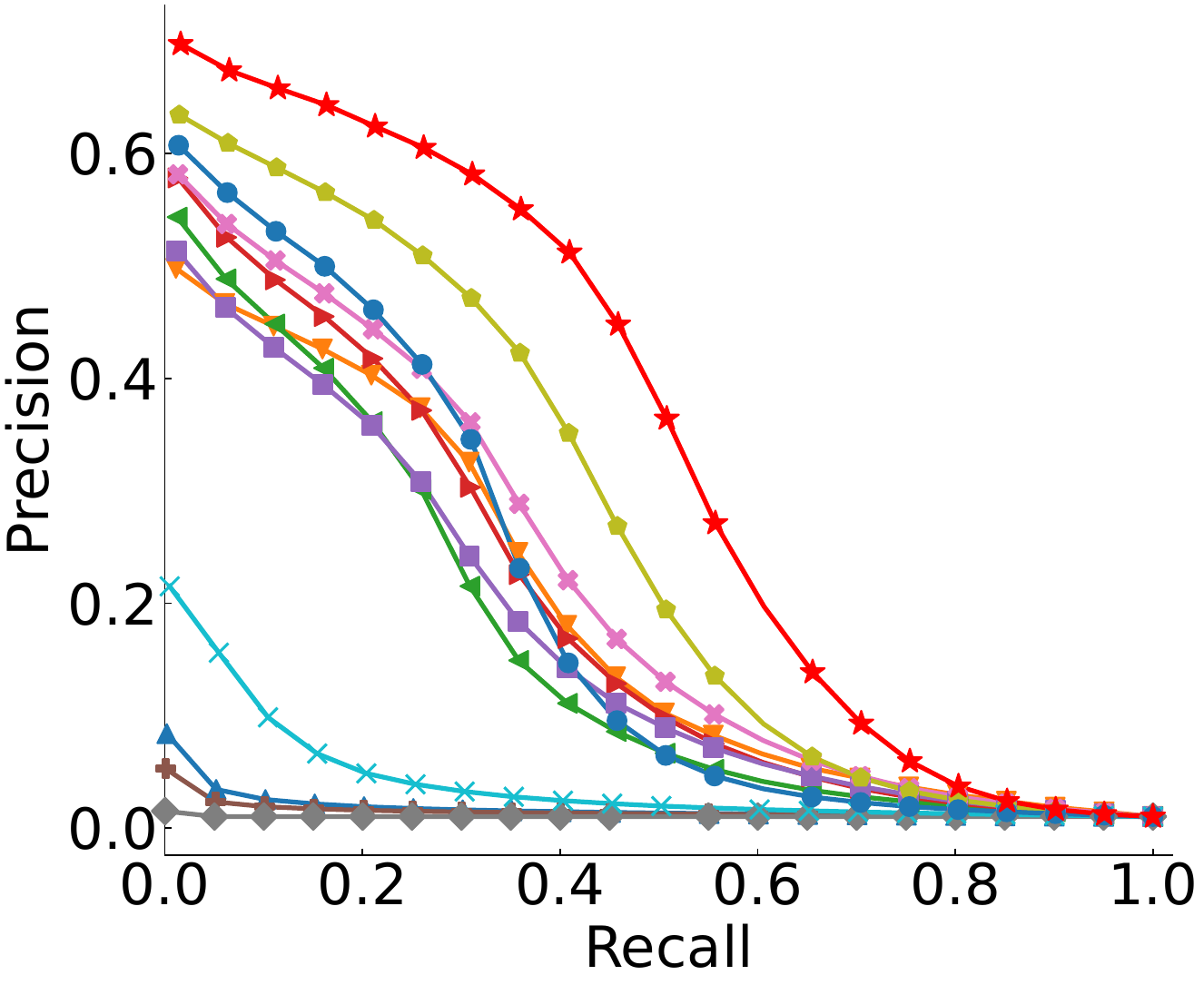} &
\includegraphics[width=0.28\textwidth]{Figures/xmedianet_T2I_0.6PR.pdf} \\
{\footnotesize (e) Wikipedia (T2I)} &
{\footnotesize (f) XMedia (T2I)} &
{\footnotesize (g) INRIA-Websearch (T2I)} &
{\footnotesize (h) XMediaNet (T2I)}
\end{tabular}
}
\caption{Precision-recall curves on four datasets under the medium missing setting with a noise ratio of 0.6.}
\label{fig:sub_pr}
\end{figure*}

\begin{table*}[!t]
\centering
{\small
\setlength{\tabcolsep}{4.5pt}
\begin{tabular}{c|c|cccccc|cccccc}
\hline
\multirow{3}{*}{Noise} &
\multirow{3}{*}{Seed} &
\multicolumn{6}{c|}{XMedia} &
\multicolumn{6}{c}{XMediaNet} \\ \cline{3-14}
& &
\multicolumn{2}{c|}{Easy} &
\multicolumn{2}{c|}{Medium} &
\multicolumn{2}{c|}{Hard} &
\multicolumn{2}{c|}{Easy} &
\multicolumn{2}{c|}{Medium} &
\multicolumn{2}{c}{Hard} \\ \cline{3-14}
& &
I2T & \multicolumn{1}{c|}{T2I} &
I2T & \multicolumn{1}{c|}{T2I} &
I2T & \multicolumn{1}{c|}{T2I} &
I2T & \multicolumn{1}{c|}{T2I} &
I2T & \multicolumn{1}{c|}{T2I} &
I2T & T2I \\ \hline

\multirow{8}{*}{0.40}
& 2021 & 90.27 & \multicolumn{1}{c|}{90.98} & 90.30 & \multicolumn{1}{c|}{91.08} & 87.60 & \multicolumn{1}{c|}{88.29} & 59.16 & \multicolumn{1}{c|}{59.06} & 55.65 & \multicolumn{1}{c|}{55.58} & 49.16 & 48.68 \\
& 2022 & 90.73 & \multicolumn{1}{c|}{91.34} & 90.02 & \multicolumn{1}{c|}{90.96} & 86.91 & \multicolumn{1}{c|}{87.60} & 59.69 & \multicolumn{1}{c|}{59.57} & 55.57 & \multicolumn{1}{c|}{55.98} & 49.06 & 48.93 \\
& 2023 & 90.87 & \multicolumn{1}{c|}{91.83} & 89.93 & \multicolumn{1}{c|}{90.71} & 87.19 & \multicolumn{1}{c|}{87.88} & 59.70 & \multicolumn{1}{c|}{59.29} & 55.81 & \multicolumn{1}{c|}{55.91} & 49.13 & 48.60 \\
& 2024 & 90.79 & \multicolumn{1}{c|}{91.11} & 90.27 & \multicolumn{1}{c|}{90.88} & 87.43 & \multicolumn{1}{c|}{88.16} & 59.23 & \multicolumn{1}{c|}{59.32} & 55.43 & \multicolumn{1}{c|}{55.79} & 49.03 & 48.71 \\
& 2025 & 90.75 & \multicolumn{1}{c|}{91.88} & 89.77 & \multicolumn{1}{c|}{90.68} & 87.68 & \multicolumn{1}{c|}{88.18} & 59.60 & \multicolumn{1}{c|}{60.00} & 56.04 & \multicolumn{1}{c|}{55.76} & 49.60 & 49.00 \\
& 2026 & 90.45 & \multicolumn{1}{c|}{91.63} & 90.12 & \multicolumn{1}{c|}{91.18} & 87.14 & \multicolumn{1}{c|}{87.71} & 59.44 & \multicolumn{1}{c|}{59.41} & 56.08 & \multicolumn{1}{c|}{55.87} & 49.05 & 48.84 \\
& $\mu$ & 90.64 & \multicolumn{1}{c|}{91.46} & 90.07 & \multicolumn{1}{c|}{90.92} & 87.33 & \multicolumn{1}{c|}{87.97} & 59.47 & \multicolumn{1}{c|}{59.44} & 55.76 & \multicolumn{1}{c|}{55.81} & 49.17 & 48.79 \\
& $\sigma$ & 0.23 & \multicolumn{1}{c|}{0.38} & 0.20 & \multicolumn{1}{c|}{0.20} & 0.30 & \multicolumn{1}{c|}{0.28} & 0.23 & \multicolumn{1}{c|}{0.32} & 0.26 & \multicolumn{1}{c|}{0.14} & 0.22 & 0.16 \\ \hline

\multirow{8}{*}{0.60}
& 2021 & 90.66 & \multicolumn{1}{c|}{91.67} & 90.47 & \multicolumn{1}{c|}{90.45} & 87.24 & \multicolumn{1}{c|}{87.81} & 59.56 & \multicolumn{1}{c|}{59.53} & 55.50 & \multicolumn{1}{c|}{55.48} & 49.38 & 48.92 \\
& 2022 & 90.59 & \multicolumn{1}{c|}{91.44} & 89.82 & \multicolumn{1}{c|}{89.98} & 87.18 & \multicolumn{1}{c|}{88.12} & 59.36 & \multicolumn{1}{c|}{59.71} & 54.85 & \multicolumn{1}{c|}{55.20} & 49.25 & 48.77 \\
& 2023 & 90.84 & \multicolumn{1}{c|}{91.45} & 89.92 & \multicolumn{1}{c|}{90.20} & 87.83 & \multicolumn{1}{c|}{87.42} & 59.09 & \multicolumn{1}{c|}{59.62} & 55.61 & \multicolumn{1}{c|}{55.64} & 49.18 & 48.61 \\
& 2024 & 90.52 & \multicolumn{1}{c|}{91.66} & 89.92 & \multicolumn{1}{c|}{90.00} & 85.70 & \multicolumn{1}{c|}{85.62} & 59.29 & \multicolumn{1}{c|}{59.87} & 55.21 & \multicolumn{1}{c|}{55.16} & 49.10 & 48.52 \\
& 2025 & 90.54 & \multicolumn{1}{c|}{91.43} & 89.68 & \multicolumn{1}{c|}{90.00} & 86.55 & \multicolumn{1}{c|}{86.77} & 59.40 & \multicolumn{1}{c|}{60.20} & 55.43 & \multicolumn{1}{c|}{55.37} & 49.20 & 48.50 \\
& 2026 & 90.77 & \multicolumn{1}{c|}{91.01} & 89.93 & \multicolumn{1}{c|}{90.90} & 85.32 & \multicolumn{1}{c|}{85.50} & 59.27 & \multicolumn{1}{c|}{59.57} & 55.53 & \multicolumn{1}{c|}{55.65} & 49.12 & 48.56 \\
& $\mu$ & 90.65 & \multicolumn{1}{c|}{91.44} & 89.96 & \multicolumn{1}{c|}{90.25} & 86.64 & \multicolumn{1}{c|}{86.87} & 59.33 & \multicolumn{1}{c|}{59.75} & 55.35 & \multicolumn{1}{c|}{55.42} & 49.21 & 48.65 \\
& $\sigma$ & 0.13 & \multicolumn{1}{c|}{0.24} & 0.27 & \multicolumn{1}{c|}{0.36} & 0.97 & \multicolumn{1}{c|}{1.11} & 0.16 & \multicolumn{1}{c|}{0.25} & 0.28 & \multicolumn{1}{c|}{0.21} & 0.10 & 0.17 \\ \hline

\end{tabular}
}
\caption{The mAP (\%) results under different random seeds on XMedia and XMediaNet, together with the mean ($\mu$) and standard deviation ($\sigma$).}
\label{tab:seed_analysis}
\end{table*}

\subsection{Implementation Details}
\label{sec:details}
\textbf{Network Architecture}: Both image and text networks adopt similar architectures. Specifically, they extract features through a two-layer multilayer perceptron (MLP). The first fully connected layer is followed by a batch normalization layer, a ReLU activation layer, and a dropout layer to prevent overfitting. The second fully connected layer is followed only by a batch normalization layer. The dimensionality changes are as follows: for the image network, $d^v$ $\rightarrow$ 1024 $\rightarrow$ 512; for the text network, $d^u$ $\rightarrow$ 1024 $\rightarrow$ 512. Following feature extraction, both networks apply $L_2$ normalization to the 512-dimensional output features.
\\
\textbf{Implementation Details}: We employ the Adam optimizer to train both the image and text networks. For the INRIA-Websearch dataset, the learning rate is linearly scheduled from $10^{-3}$ to $10^{-4}$ over the course of training, while for the Wikipedia, XMedia, and XMediaNet datasets, the learning rate is maintained at a constant $10^{-4}$. The training batch size is set to 128 for the Wikipedia dataset and 256 for the other datasets. The maximum number of training epochs is set to 100. Our SPIRN is implemented in PyTorch and trained on a machine equipped with one NVIDIA RTX 4090 24GB GPU, a 13-core CPU, and 56 GB RAM. The software environment uses Python 3.8.10 and PyTorch 2.0.1+cu118, along with torchvision 0.15.2+cu118 and torchaudio 2.0.2+cu118.
\\
\textbf{Incomplete Modality Scenarios}: 
Following the settings of previous incomplete cross-modal retrieval works, we design three settings of increasing severity, denoted as easy, medium, and hard, with data distribution ratios of (80\%, 10\%, 10\%), (50\%, 25\%, 25\%), and (20\%, 40\%, 40\%), respectively. These ratios represent the proportions of samples with complete paired modalities, image-only modalities, and text-only modalities in the training set.
\\
\textbf{Noisy Label Scenarios}: 
To evaluate the robustness of SPIRN to label noise, we inject synthetic noise into the training set while keeping the validation and test sets clean. Under the symmetric random-label noise setting, we randomly select 20\%, 40\%, 60\%, or 80\% of the training samples and uniformly resample their labels.
\\
\textbf{Evaluation Metrics}: In our experiments, we adopt mean average precision (mAP) \cite{cui2019scalable,li2019relaxed} as the primary evaluation metric, because it summarizes ranking quality over the entire retrieval list and is widely used for cross-modal retrieval. The retrieval performance is assessed on two tasks: image-to-text retrieval (I2T) and text-to-image retrieval (T2I). Moreover, we further employ precision-recall (PR) curves to provide a more comprehensive view of performance across different recall levels, which is especially informative under incomplete and noisy settings.

\begin{table*}[!t]
\centering
{\small
\setlength{\tabcolsep}{4.5pt}
\begin{tabular}{c|c|cccccccccccc}
\hline
\multirow{3}{*}{Dataset} & \multirow{3}{*}{Method} &
\multicolumn{12}{c}{Noise Ratio} \\ \cline{3-14}
& & \multicolumn{3}{c|}{0.2} & \multicolumn{3}{c|}{0.4} &
\multicolumn{3}{c|}{0.6} & \multicolumn{3}{c}{0.8} \\ \cline{3-14}
& & I2T & T2I & \multicolumn{1}{c|}{Mean} &
I2T & T2I & \multicolumn{1}{c|}{Mean} &
I2T & T2I & \multicolumn{1}{c|}{Mean} &
I2T & T2I & Mean \\ \hline
\multirow{12}{*}{XMedia}    & ALGCN                   & 47.1          & 15.0          & \multicolumn{1}{c|}{31.0}          & 29.5          & 11.9          & \multicolumn{1}{c|}{20.7}          & 13.1          & 8.5           & \multicolumn{1}{c|}{10.8}          & 8.5           & 7.2           & 7.8           \\
& UCCH                    & 55.5          & 59.9          & \multicolumn{1}{c|}{57.7}          & 55.5          & 59.9          & \multicolumn{1}{c|}{57.7}          & 55.5          & 59.9          & \multicolumn{1}{c|}{57.7}          & 55.5          & 59.9          & 57.7          \\
& GNN4CMR                 & 57.7          & 61.9          & \multicolumn{1}{c|}{59.8}          & 48.1          & 52.4          & \multicolumn{1}{c|}{50.2}          & 36.3          & 38.3          & \multicolumn{1}{c|}{37.3}          & 28.5          & 28.5          & 28.5          \\
& RONO                    & 53.6          & 52.0          & \multicolumn{1}{c|}{52.8}          & 48.7          & 46.7          & \multicolumn{1}{c|}{47.7}          & 40.0          & 38.4          & \multicolumn{1}{c|}{39.2}          & 35.3          & 33.8          & 34.5          \\
& CIRH                    & 39.0          & 35.4          & \multicolumn{1}{c|}{37.2}          & 39.0          & 35.4          & \multicolumn{1}{c|}{37.2}          & 39.0          & 35.4          & \multicolumn{1}{c|}{37.2}          & 39.0          & 35.4          & 37.2          \\
& NCH                     & 9.0           & 8.0           & \multicolumn{1}{c|}{8.5}           & 6.7           & 8.1           & \multicolumn{1}{c|}{7.4}           & 7.5           & 8.5           & \multicolumn{1}{c|}{8.0}           & 7.5           & 8.8           & 8.1           \\
& NRCH                    & 59.6          & 60.0          & \multicolumn{1}{c|}{59.8}          & 58.6          & 58.1          & \multicolumn{1}{c|}{58.4}          & 53.7          & 53.2          & \multicolumn{1}{c|}{53.5}          & 53.8          & 53.4          & 53.6          \\
& CICH                    & 6.3           & 7.3           & \multicolumn{1}{c|}{6.8}           & 6.3           & 8.5           & \multicolumn{1}{c|}{7.4}           & 6.1           & 6.9           & \multicolumn{1}{c|}{6.5}           & 6.3           & 6.3           & 6.3           \\
& RSHNL & \underline{78.6} & \underline{78.4} & \multicolumn{1}{c|}{\underline{78.5}} &
\underline{72.4} & \underline{72.1} & \multicolumn{1}{c|}{\underline{72.2}} &
\underline{63.7} & \underline{62.7} & \multicolumn{1}{c|}{\underline{63.2}} &
\underline{47.9} & \underline{48.0} & \underline{47.9} \\
& DHRL                    & 69.6          & 67.8          & \multicolumn{1}{c|}{68.7}          & 46.2          & 48.0          & \multicolumn{1}{c|}{47.1}          & 21.9          & 22.7          & \multicolumn{1}{c|}{22.3}          & 13.2          & 13.0          & 13.1          \\
& EGICH                   & 77.1          & 77.0          & \multicolumn{1}{c|}{77.1}          & 53.5          & 53.5          & \multicolumn{1}{c|}{53.5}          & 25.5          & 25.8          & \multicolumn{1}{c|}{25.7}          & 9.6           & 9.9           & 9.8           \\
& SPIRN                   & \textbf{87.8} & \textbf{88.6} & \multicolumn{1}{c|}{\textbf{88.2}} & \textbf{86.2} & \textbf{86.7} & \multicolumn{1}{c|}{\textbf{86.5}} & \textbf{85.8} & \textbf{86.6} & \multicolumn{1}{c|}{\textbf{86.2}} & \textbf{76.7} & \textbf{75.5} & \textbf{76.1} \\ \hline
\multirow{12}{*}{XMediaNet} & ALGCN                   & 0.8           & 1.0           & \multicolumn{1}{c|}{0.9}           & 0.8           & 0.9           & \multicolumn{1}{c|}{0.9}           & 0.8           & 0.9           & \multicolumn{1}{c|}{0.8}           & 0.8           & 0.9           & 0.8           \\
& UCCH                    & 7.3           & 9.5           & \multicolumn{1}{c|}{8.4}           & 7.3           & 9.5           & \multicolumn{1}{c|}{8.4}           & 7.3           & 9.5           & \multicolumn{1}{c|}{8.4}           & 7.3           & 9.5           & 8.4           \\
& GNN4CMR & \underline{25.7} & \underline{28.9} & \multicolumn{1}{c|}{\underline{27.3}} &
\underline{19.6} & \underline{22.4} & \multicolumn{1}{c|}{\underline{21.0}} &
\underline{15.1} & \underline{17.5} & \multicolumn{1}{c|}{\underline{16.3}} &
\underline{14.8} & \underline{16.7} & \underline{15.7} \\
& RONO                    & 9.2           & 9.4           & \multicolumn{1}{c|}{9.3}           & 9.9           & 10.3          & \multicolumn{1}{c|}{10.1}          & 9.5           & 9.7           & \multicolumn{1}{c|}{9.6}           & 8.7           & 9.1           & 8.9           \\
& CIRH                    & 3.5           & 3.7           & \multicolumn{1}{c|}{3.6}           & 3.5           & 3.7           & \multicolumn{1}{c|}{3.6}           & 3.5           & 3.7           & \multicolumn{1}{c|}{3.6}           & 3.5           & 3.7           & 3.6           \\
& NCH                     & 0.8           & 1.1           & \multicolumn{1}{c|}{1.0}           & 0.8           & 1.1           & \multicolumn{1}{c|}{0.9}           & 0.7           & 1.0           & \multicolumn{1}{c|}{0.9}           & 0.8           & 1.0           & 0.9           \\
& NRCH                    & 6.4           & 7.4           & \multicolumn{1}{c|}{6.9}           & 6.5           & 7.6           & \multicolumn{1}{c|}{7.0}           & 6.6           & 7.6           & \multicolumn{1}{c|}{7.1}           & 6.3           & 7.3           & 6.8           \\
& CICH                    & 1.3           & 2.2           & \multicolumn{1}{c|}{1.7}           & 1.1           & 1.4           & \multicolumn{1}{c|}{1.2}           & 0.9           & 1.0           & \multicolumn{1}{c|}{0.9}           & 0.7           & 0.8           & 0.8           \\
& RSHNL                   & 12.5          & 13.7          & \multicolumn{1}{c|}{13.1}          & 8.9           & 9.9           & \multicolumn{1}{c|}{9.4}           & 6.9           & 7.8           & \multicolumn{1}{c|}{7.3}           & 5.8           & 6.5           & 6.1           \\
& DHRL                    & 5.4           & 6.4           & \multicolumn{1}{c|}{5.9}           & 3.0           & 3.7           & \multicolumn{1}{c|}{3.4}           & 1.5           & 1.5           & \multicolumn{1}{c|}{1.5}           & 0.9           & 1.0           & 1.0           \\
& EGICH                   & 17.1          & 19.6          & \multicolumn{1}{c|}{18.3}          & 12.0          & 13.9          & \multicolumn{1}{c|}{13.0}          & 7.9           & 10.0          & \multicolumn{1}{c|}{8.9}           & 6.0           & 7.1           & 6.5           \\
& SPIRN                   & \textbf{45.5} & \textbf{41.5} & \multicolumn{1}{c|}{\textbf{43.5}} & \textbf{44.7} & \textbf{42.8} & \multicolumn{1}{c|}{\textbf{43.8}} & \textbf{45.4} & \textbf{44.0} & \multicolumn{1}{c|}{\textbf{44.7}} & \textbf{33.2} & \textbf{33.2} & \textbf{33.2} \\ \hline
\end{tabular}
}
\caption{Comparison of mAP (\%) on XMedia and XMediaNet under a 90\% missing-modality setting and four noise ratios.}
\label{tab:severe_missingness}
\end{table*}

\subsection{Comparison Methods}
\label{sec:methods}
To verify the superiority of our proposed SPIRN, we extensively compare SPIRN with 11 state-of-the-art baselines, as shown in Tab. \ref{tab:comparison_methods}. These include two unsupervised methods (i.e., UCCH \cite{UCCH} and CIRH \cite{CIRH}) and nine supervised methods (i.e., ALGCN \cite{ALGCN}, GNN4CMR \cite{GNN4CMR}, RONO \cite{feng2023rono}, NCH \cite{NCH}, NRCH \cite{NRCH}, CICH \cite{luo2024contrastive}, RSHNL \cite{RSHNL}, DHRL \cite{DHRL}, and EGICH \cite{EGICH}). 
Notably, RONO, NRCH, RSHNL, and DHRL are specifically designed to tackle noisy-label learning, whereas GNN4CMR, NCH, CICH, and EGICH are specifically tailored for missing-modality learning. Regarding output representations, we evaluate each baseline using its native representation and retrieval protocol. ALGCN, GNN4CMR, and RONO are continuous methods that use real-valued embeddings, while the remaining baselines are hashing methods that use 128-bit hash codes. SPIRN is evaluated as a continuous real-valued method using 512-dimensional L2-normalized embeddings with cosine-similarity ranking.
To ensure fairness, we rerun the code for all SOTA methods using the parameters provided in their papers or released code and freeze the feature extractor during training. We use validation mAP to select the optimal model weights and report the final mAP scores on the test set. 
In all experimental tables, the best results are denoted in \textbf{bold}, while the second-best results are \underline{underlined}.

\begin{figure}[!t]
\centering
\includegraphics[width=0.9\columnwidth]{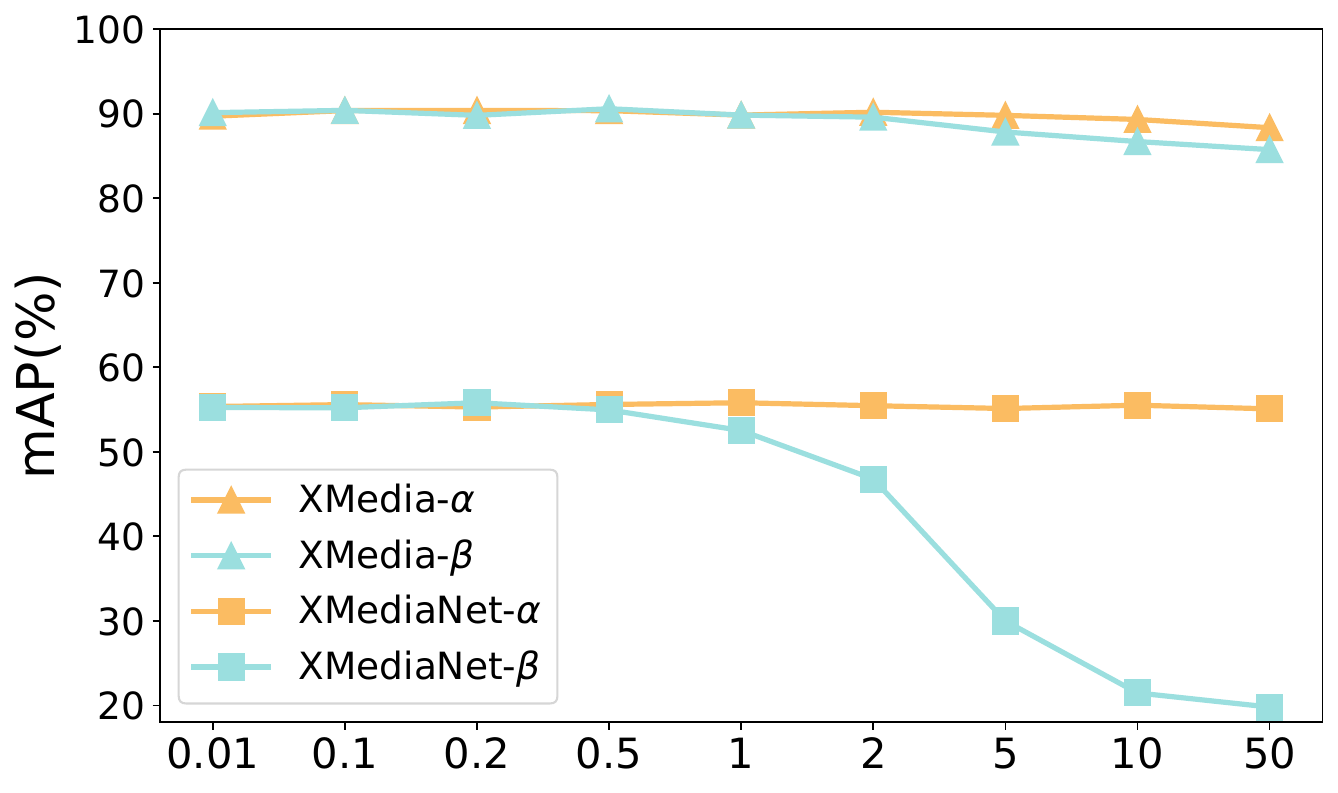}\\
{\footnotesize (a) Different $\alpha$ and $\beta$}\\[2pt]
\includegraphics[width=0.9\columnwidth]{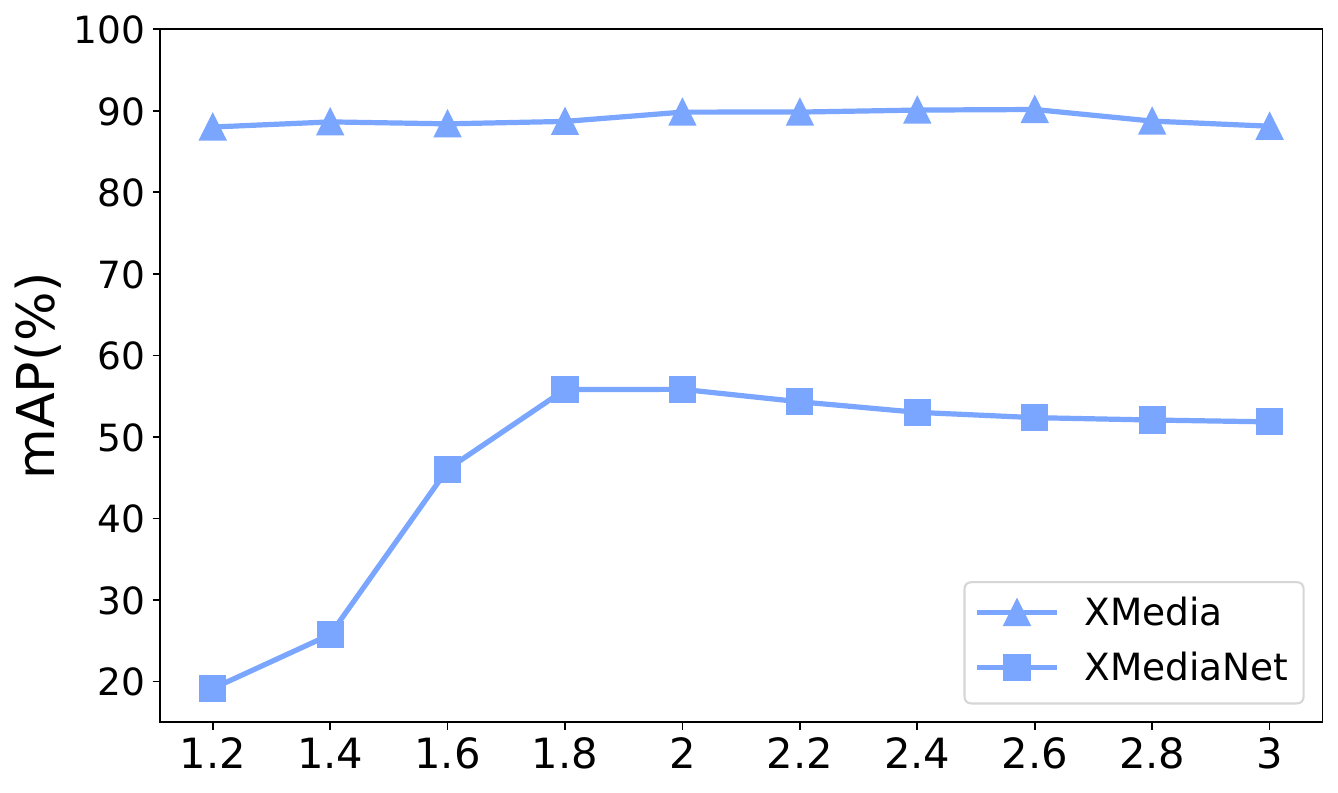}\\
{\footnotesize (b) Different $\eta$}
\caption{The mAP (\%) results under the medium missing scene for different $\alpha$, $\beta$, and $\eta$.}
\label{fig:parameter}
\end{figure}

\begin{table*}[!t]
\centering
{\small
\begin{tabular}{c|l|cccccccccccc}
\hline
\multirow{3}{*}{Dataset}    & \multicolumn{1}{c|}{\multirow{3}{*}{Method}} & \multicolumn{12}{c}{Noise Ratio}                                 \\ \cline{3-14} 
& \multicolumn{1}{c|}{}                        & \multicolumn{3}{c|}{0.1}                                           & \multicolumn{3}{c|}{0.2}                                           & \multicolumn{3}{c|}{0.3}                                           & \multicolumn{3}{c}{0.4}                       \\ \cline{3-14} 
& \multicolumn{1}{c|}{}                        & I2T           & T2I           & \multicolumn{1}{c|}{Mean}          & I2T           & T2I           & \multicolumn{1}{c|}{Mean}          & I2T           & T2I           & \multicolumn{1}{c|}{Mean}          & I2T           & T2I           & Mean          \\ \hline
\multirow{12}{*}{XMedia}    & ALGCN                                        & 85.6          & 84.4          & \multicolumn{1}{c|}{85.0}          & 82.4          & 79.8          & \multicolumn{1}{c|}{81.1}          & 75.3          & 70.3          & \multicolumn{1}{c|}{72.8}          & 65.0          & 56.9          & 61.0          \\
& UCCH                                         & 72.7          & 74.4          & \multicolumn{1}{c|}{73.5}          & 72.7          & 74.4          & \multicolumn{1}{c|}{73.5}          & 72.7          & 74.4          & \multicolumn{1}{c|}{73.5}          & 72.7          & 74.4          & 73.5          \\
& GNN4CMR                                      & 84.7          & 85.3          & \multicolumn{1}{c|}{85.0}          & 82.9          & 81.9          & \multicolumn{1}{c|}{82.4}          & 78.7          & 78.8          & \multicolumn{1}{c|}{78.7}          & 76.1          & 76.5          & 76.3          \\
& RONO                                         & {\underline{89.1}}    & {\underline{90.0}}    & \multicolumn{1}{c|}{{\underline {89.6}}}    & {\underline{89.3}}    & {\underline{90.1}}    & \multicolumn{1}{c|}{{\underline{89.7}}}    & {\underline{86.9}}    & {\underline{88.1}}    & \multicolumn{1}{c|}{{\underline{87.5}}}    & 86.4          & 86.5          & 86.4          \\
& CIRH                                         & 62.9          & 60.2          & \multicolumn{1}{c|}{61.5}          & 62.9          & 60.2          & \multicolumn{1}{c|}{61.5}          & 62.9          & 60.2          & \multicolumn{1}{c|}{61.5}          & 62.9          & 60.2          & 61.5          \\
& NCH                                          & 11.0          & 8.9           & \multicolumn{1}{c|}{9.9}           & 10.8          & 8.5           & \multicolumn{1}{c|}{9.7}           & 8.2           & 8.9           & \multicolumn{1}{c|}{8.6}           & 7.8           & 8.2           & 8.0           \\
& NRCH                                         & 85.2          & 87.4          & \multicolumn{1}{c|}{86.3}          & 84.8          & 85.4          & \multicolumn{1}{c|}{85.1}          & 83.0          & 83.9          & \multicolumn{1}{c|}{83.5}          & 80.5          & 81.8          & 81.2          \\
& CICH                                         & 7.3           & 6.7           & \multicolumn{1}{c|}{7.0}           & 7.0           & 7.4           & \multicolumn{1}{c|}{7.2}           & 6.1           & 6.9           & \multicolumn{1}{c|}{6.5}           & 6.2           & 7.6           & 6.9           \\
& RSHNL                                        & 88.5          & 88.9          & \multicolumn{1}{c|}{88.7}          & 88.5          & 89.0          & \multicolumn{1}{c|}{88.8}          & 86.4          & 86.7          & \multicolumn{1}{c|}{86.5}          & {\underline{87.4}}    & {\underline{86.8}}    & {\underline{87.1}}    \\
& DHRL                                         & 87.8          & 88.0          & \multicolumn{1}{c|}{87.9}          & 88.7          & 88.4          & \multicolumn{1}{c|}{88.5}          & 84.5          & 84.9          & \multicolumn{1}{c|}{84.7}          & 78.1          & 77.0          & 77.5          \\
& EGICH                                        & 86.7          & 87.3          & \multicolumn{1}{c|}{87.0}          & 83.7          & 83.3          & \multicolumn{1}{c|}{83.5}          & 75.3          & 75.8          & \multicolumn{1}{c|}{75.6}          & 66.5          & 66.3          & 66.4          \\
& SPIRN                                        & \textbf{89.5} & \textbf{91.1} & \multicolumn{1}{c|}{\textbf{90.3}} & \textbf{89.7} & \textbf{90.6} & \multicolumn{1}{c|}{\textbf{90.1}} & \textbf{90.0} & \textbf{90.7} & \multicolumn{1}{c|}{\textbf{90.3}} & \textbf{89.9} & \textbf{90.7} & \textbf{90.3} \\ \hline
\multirow{12}{*}{XMediaNet} & ALGCN                                        & 18.0          & 18.7          & \multicolumn{1}{c|}{18.4}          & 11.9          & 11.1          & \multicolumn{1}{c|}{11.5}          & 6.8           & 5.6           & \multicolumn{1}{c|}{6.2}           & 4.0           & 3.3           & 3.6           \\
& UCCH                                         & 15.7          & 17.4          & \multicolumn{1}{c|}{16.6}          & 15.7          & 17.4          & \multicolumn{1}{c|}{16.6}          & 15.7          & 17.4          & \multicolumn{1}{c|}{16.6}          & 15.7          & 17.4          & 16.6          \\
& GNN4CMR                                      & {\underline{50.2}}    & {\underline{50.6}}    & \multicolumn{1}{c|}{{\underline{50.4}}}    & {\underline{47.6}}    & {\underline{48.3}}    & \multicolumn{1}{c|}{{\underline{47.9}}}    & {\underline{44.4}}    & 44.9          & \multicolumn{1}{c|}{44.6}          & 40.8          & 41.8          & 41.3          \\
& RONO                                         & 16.5          & 13.9          & \multicolumn{1}{c|}{15.2}          & 15.2          & 12.9          & \multicolumn{1}{c|}{14.1}          & 14.5          & 12.6          & \multicolumn{1}{c|}{13.5}          & 14.0          & 12.2          & 13.1          \\
& CIRH                                         & 11.5          & 11.2          & \multicolumn{1}{c|}{11.3}          & 11.5          & 11.2          & \multicolumn{1}{c|}{11.3}          & 11.5          & 11.2          & \multicolumn{1}{c|}{11.3}          & 11.5          & 11.2          & 11.3          \\
& NCH                                          & 0.9           & 1.2           & \multicolumn{1}{c|}{1.0}           & 0.9           & 1.3           & \multicolumn{1}{c|}{1.1}           & 0.8           & 1.1           & \multicolumn{1}{c|}{0.9}           & 0.8           & 1.1           & 0.9           \\
& NRCH                                         & 14.8          & 17.1          & \multicolumn{1}{c|}{15.9}          & 14.5          & 16.8          & \multicolumn{1}{c|}{15.7}          & 14.9          & 17.1          & \multicolumn{1}{c|}{16.0}          & 14.7          & 17.1          & 15.9          \\
& CICH                                         & 1.5           & 2.6           & \multicolumn{1}{c|}{2.1}           & 1.7           & 2.3           & \multicolumn{1}{c|}{2.0}           & 1.9           & 2.0           & \multicolumn{1}{c|}{1.9}           & 1.4           & 1.4           & 1.4           \\
& RSHNL                                        & 38.8          & 41.1          & \multicolumn{1}{c|}{40.0}          & 38.5          & 40.9          & \multicolumn{1}{c|}{39.7}          & 38.7          & 40.7          & \multicolumn{1}{c|}{39.7}          & 37.8          & 39.7          & 38.7          \\
& DHRL                                         & 20.1          & 24.1          & \multicolumn{1}{c|}{22.1}          & 14.0          & 17.2          & \multicolumn{1}{c|}{15.6}          & 11.5          & 14.1          & \multicolumn{1}{c|}{12.8}          & 7.9           & 10.2          & 9.0           \\
& EGICH                                        & 46.0          & 47.9          & \multicolumn{1}{c|}{46.9}          & 45.0          & 46.9          & \multicolumn{1}{c|}{46.0}          & 44.0          & {\underline{45.8}}    & \multicolumn{1}{c|}{{\underline{44.9}}}    & {\underline{41.5}}    & {\underline{43.3}}    & {\underline{42.4}}    \\
& SPIRN                                        & \textbf{55.8} & \textbf{56.3} & \multicolumn{1}{c|}{\textbf{56.0}} & \textbf{55.8} & \textbf{56.2} & \multicolumn{1}{c|}{\textbf{56.0}} & \textbf{55.8} & \textbf{55.6} & \multicolumn{1}{c|}{\textbf{55.7}} & \textbf{55.9} & \textbf{56.2} & \textbf{56.0} \\ \hline
\end{tabular}
}
\caption{Comparison of mAP (\%) on XMedia and XMediaNet under asymmetric label noise with varying noise ratios in the medium missing setting.}
\label{tab:asym_noise}
\end{table*}

\subsection{Additional Experiments}
\subsubsection{Qualitative Comparisons}
\label{sec:pr}
To comprehensively evaluate the retrieval performance of SPIRN against several state-of-the-art methods, we present the Precision-Recall (PR) curves under the medium missing-data scenario with a 0.6 label noise ratio in Fig. \ref{fig:sub_pr}. As evidenced by these results, SPIRN consistently maintains higher precision across all recall levels, further demonstrating its effectiveness and superiority.

\subsubsection{Seed Analysis}
To assess the robustness of SPIRN to random initialization, we conduct experiments using six random seeds under different noise ratios and modality-missing settings on XMedia and XMediaNet. Tab.~\ref{tab:seed_analysis} reports the complete mAP results for both image-to-text (I2T) and text-to-image (T2I) retrieval, together with the mean $\mu$ and standard deviation $\sigma$ across seeds. 
SPIRN consistently achieves competitive performance across all seeds and settings, with small standard deviations on both datasets. These results indicate that the observed performance gains are not attributable to a favorable random seed, while further demonstrating the stability and reliability of SPIRN under coupled modality incompleteness and label noise.

\subsubsection{Severe Missingness}
To further evaluate robustness under extreme modality incompleteness, we consider a severe missing setting in which 90\% of the training instances are incomplete, leaving only 10\% complete image-text pairs. This setting substantially weakens the available cross-modal correspondences and makes reliable label correction and missing-modality recovery particularly challenging. As shown in Tab.~\ref{tab:severe_missingness}, SPIRN consistently achieves the best performance on both XMedia and XMediaNet across all four noise ratios.
Compared with the strongest competing method, SPIRN improves the mean mAP by 9.7, 14.3, 23.0, and 28.2 percentage points on XMedia, and by 16.2, 22.8, 28.4, and 17.5 percentage points on XMediaNet, under noise ratios of 0.2, 0.4, 0.6, and 0.8, respectively. These results demonstrate the robustness of SPIRN under severe modality missingness and label corruption.

\subsubsection{Parameter Analysis}
\label{sec:parameter}
In our training objective, $\alpha$ and $\beta$ control the reverse learning loss and the contrastive learning loss, respectively, whereas $\eta$ controls the tolerance to label noise.
Except for $\beta$ and $\eta$, all other hyperparameters remain consistent across datasets. To investigate their sensitivity, we conduct experiments under a label noise ratio of 0.6 and the medium missing setting. We evaluate performance across various combinations of $\alpha$, $\beta$, and $\eta$. As shown in Fig.~\ref{fig:parameter}, the results indicate the following:
(1) The model performance remains highly stable across the entire experimental range of $\alpha$, demonstrating strong robustness. For $\beta$, the performance is stable on XMedia but drops significantly on XMediaNet at larger values. This divergence stems from the semantic scale: XMedia contains only 20 categories with a relatively simple semantic space, where aggressive instance-level contrastive alignment consistently yields positive gains. However, XMediaNet is a fine-grained dataset with 200 categories. An excessively large $\beta$ causes the model to overemphasize instance-level pushing and pulling, which sacrifices category-level semantic clustering and disrupts subtle fine-grained semantic boundaries.
(2) The performance curve of $\eta$ exhibits a distinct pattern of initially rising and then declining. An overly small $\eta$ imposes a strict threshold that admits too few clean samples, leading to data under-utilization and restricting feature representation capacity. Conversely, an excessively large $\eta$ sets a loose threshold that fails to filter out severe label noise, causing the model to overfit to corrupted labels.
(3) Across all parameter variations, XMediaNet consistently exhibits more dramatic fluctuations than XMedia. This is to be expected, as optimizing a robust cross-modal common space over a denser category distribution is inherently a far more challenging optimization problem, particularly when confronted with the dual challenges of severe noise and missing modalities.
(4) Overall, the proposed SPIRN achieves highly competitive and stable performance across a broad parameter range, specifically when $\alpha, \beta \in [0.01, 0.5]$ and $\eta \in [1.8, 2.2]$.

\begin{figure}[!t]
\centering
\includegraphics[width=0.9\columnwidth]{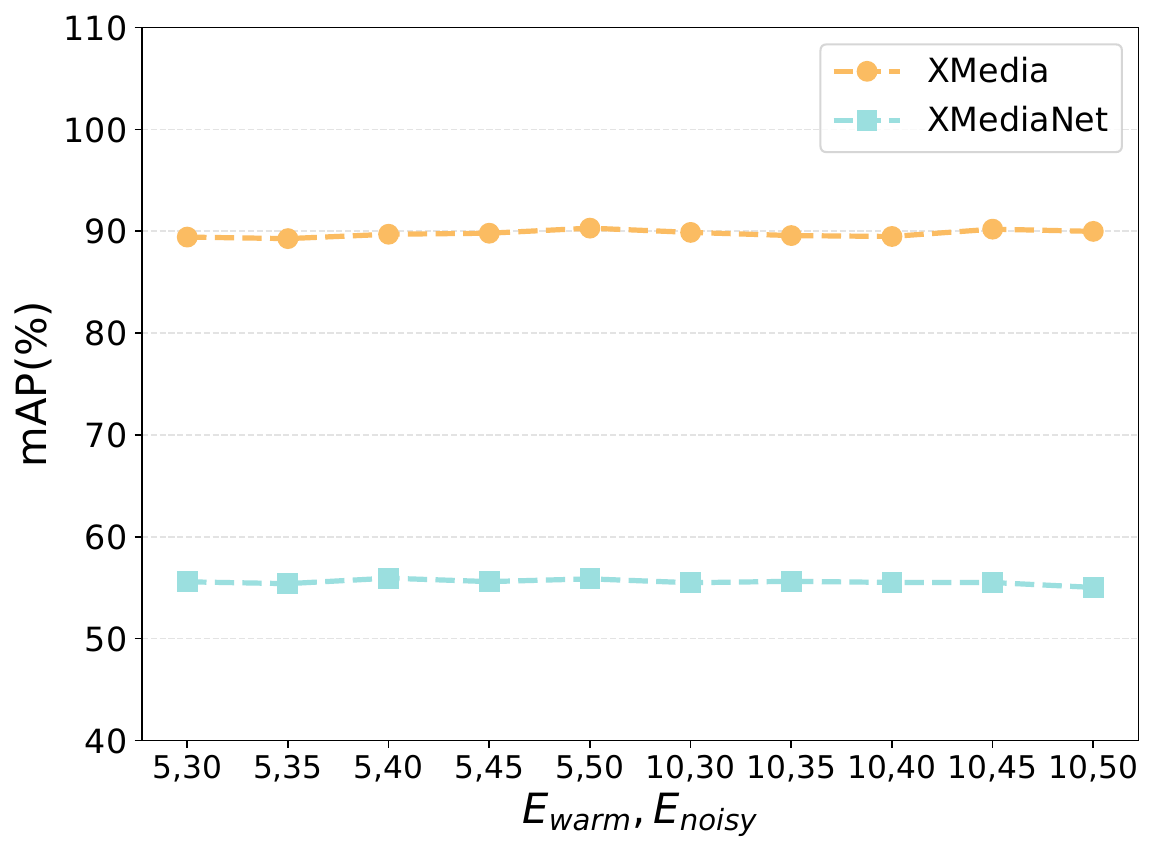}\\
\caption{The mAP (\%) results of SPIRN under different epoch combinations.}
\label{fig:epoch}
\end{figure}

\begin{figure}[!t]
\centering
\includegraphics[width=0.9\columnwidth]{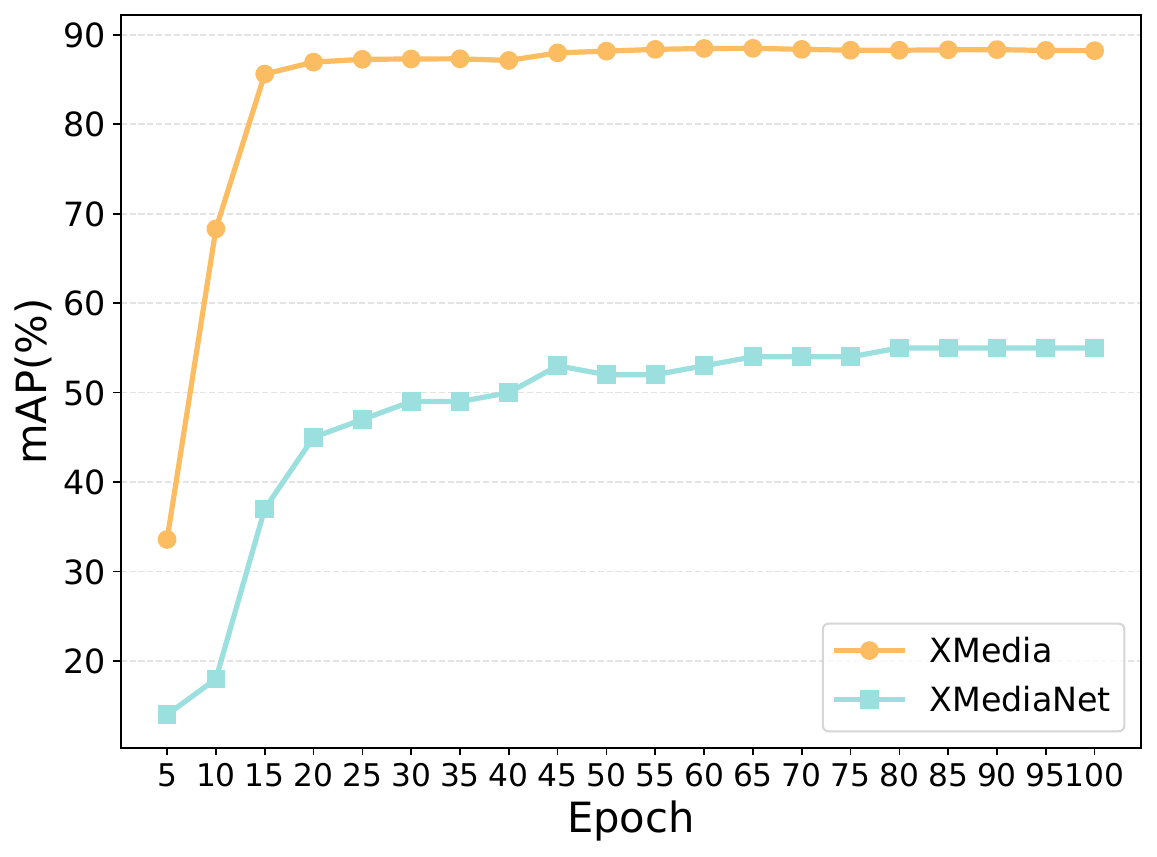}
\caption{The mAP curves of SPIRN under the medium missing modality setting with a label noise ratio of 0.6.}
\label{fig:map_line}
\end{figure}

\begin{figure}[!t]
\centering
\includegraphics[width=0.9\columnwidth]{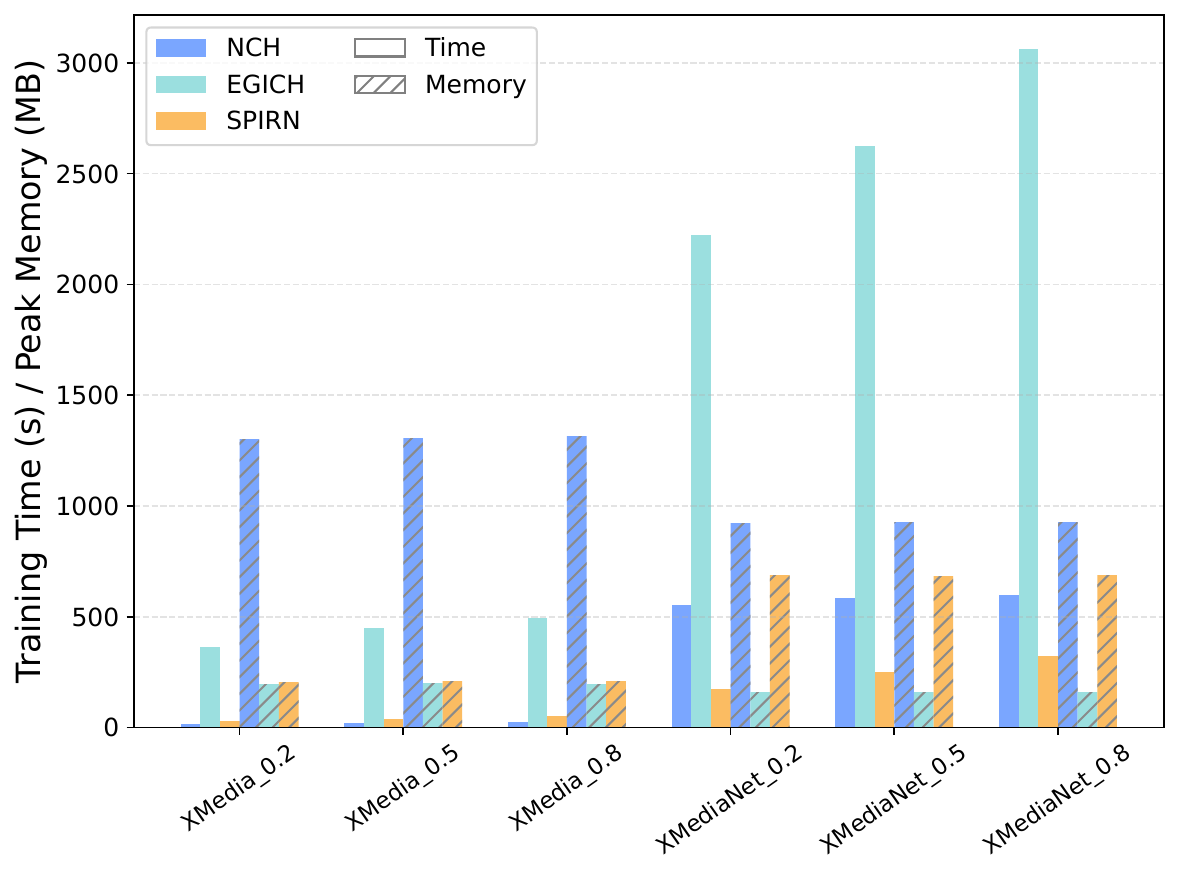}\\
\caption{Training time (s) and GPU memory consumption (MB) of different methods under varying modality-completeness settings, averaged over four label-noise ratios.}
\label{fig:time_memory}
\end{figure}

\subsubsection{Asymmetric Label Noise Analysis}
To further emulate practical annotation errors that are not uniformly distributed across categories, we evaluate SPIRN under asymmetric label noise. We adopt the medium missing setting and compare all methods under noise ratios ranging from 0.1 to 0.4 on XMedia and XMediaNet. As reported in Table~\ref{tab:asym_noise}, SPIRN consistently achieves the best mAP results across both retrieval directions, datasets, and noise ratios. These results further demonstrate that SPIRN remains robust to class-dependent label corruption, benefiting from stable semantic-prior guidance for reliable representation learning and label calibration under modality incompleteness.

\subsubsection{Epoch Sensitivity Analysis}
Since SPIRN consists of three progressive training stages, the transition epochs $E_{\mathrm{warm}}$ and $E_{\mathrm{noisy}}$ may affect its performance. 
To examine the sensitivity of SPIRN to these epoch settings, we construct ten combinations of $E_{\mathrm{warm}}$ and $E_{\mathrm{noisy}}$, while fixing the total number of training epochs to 100.
As illustrated in Fig.~\ref{fig:epoch}, SPIRN exhibits consistently stable retrieval performance across the ten epoch configurations on both XMedia and XMediaNet. This result demonstrates that SPIRN is not sensitive to precisely chosen transition epochs and does not rely on fine-grained epoch tuning. Such robustness can be attributed to the semantic prior-guided framework, which uses fixed semantic priors as consistent references throughout robust representation learning, noisy-label handling, and missing-modality completion. This design stabilizes the optimization process and facilitates smooth transitions between the three training stages. Notably, the epoch configurations are applied to all four datasets, further demonstrating the generality and practical robustness of SPIRN.

\subsubsection{mAP Curves}
To further examine the training stability of SPIRN, we plot the mAP curves under the medium missing modality setting with a label noise ratio of 0.6, as shown in Fig.~\ref{fig:map_line}. Overall, the mAP values remain stable throughout the training process and exhibit a gradual upward trend as optimization proceeds. In particular, no substantial performance degradation is observed at the transitions between the three training stages, indicating that the proposed progressive learning strategy provides a smooth and stable optimization trajectory. These results further demonstrate that SPIRN can effectively learn robust cross-modal representations despite the simultaneous presence of noisy labels and missing modalities.

\begin{table*}[!t]
\centering
{\small
\setlength{\tabcolsep}{4.5pt}
\begin{tabular}{c|c|cccccccccccc}
\hline
\multirow{3}{*}{Dataset} &
\multirow{3}{*}{Method} &
\multicolumn{12}{c}{Noise Ratio} \\ \cline{3-14}
& &
\multicolumn{3}{c|}{0.2} &
\multicolumn{3}{c|}{0.4} &
\multicolumn{3}{c|}{0.6} &
\multicolumn{3}{c}{0.8} \\ \cline{3-14}
& &
I2T & T2I & \multicolumn{1}{c|}{Mean} &
I2T & T2I & \multicolumn{1}{c|}{Mean} &
I2T & T2I & \multicolumn{1}{c|}{Mean} &
I2T & T2I & Mean \\ \hline

\multirow{5}{*}{XMediaNet}
& SPIRN-M & 33.4 & 36.5 & \multicolumn{1}{c|}{35.0}
& 30.8 & 33.4 & \multicolumn{1}{c|}{32.1}
& 24.4 & 26.5 & \multicolumn{1}{c|}{25.5}
& 15.6 & 18.0 & 16.8 \\

& SPIRN-E & 31.8 & 34.1 & \multicolumn{1}{c|}{32.9}
& 27.3 & 29.4 & \multicolumn{1}{c|}{28.4}
& 20.5 & 22.0 & \multicolumn{1}{c|}{21.2}
& 18.3 & 20.6 & 19.4 \\

& SPIRN-L & 40.9 & 43.8 & \multicolumn{1}{c|}{42.3}
& 40.0 & 42.9 & \multicolumn{1}{c|}{41.5}
& 37.2 & 39.9 & \multicolumn{1}{c|}{38.5}
& 26.5 & 27.9 & 27.2 \\

& SPIRN-R & 55.4 & 55.4 & \multicolumn{1}{c|}{55.4}
& 55.2 & 55.2 & \multicolumn{1}{c|}{55.2}
& 55.5 & 54.6 & \multicolumn{1}{c|}{55.0}
& 52.9 & 51.8 & 52.3 \\

& SPIRN & \textbf{56.0} & \textbf{56.8} &
\multicolumn{1}{c|}{\textbf{56.4}}
& \textbf{55.9} & \textbf{56.2} &
\multicolumn{1}{c|}{\textbf{56.1}}
& \textbf{55.8} & \textbf{55.8} &
\multicolumn{1}{c|}{\textbf{55.8}}
& \textbf{53.7} & \textbf{53.2} & \textbf{53.5} \\ \hline

\end{tabular}
}
\caption{Comparison of different semantic-prior constructions on XMediaNet under the medium missing modality setting and four noise ratios.}
\label{tab:prior_construction}
\end{table*}

\begin{figure*}[!t]
\centering
{
\setlength{\tabcolsep}{3pt}
\begin{tabular}{ccccc}
\includegraphics[width=0.18\textwidth]{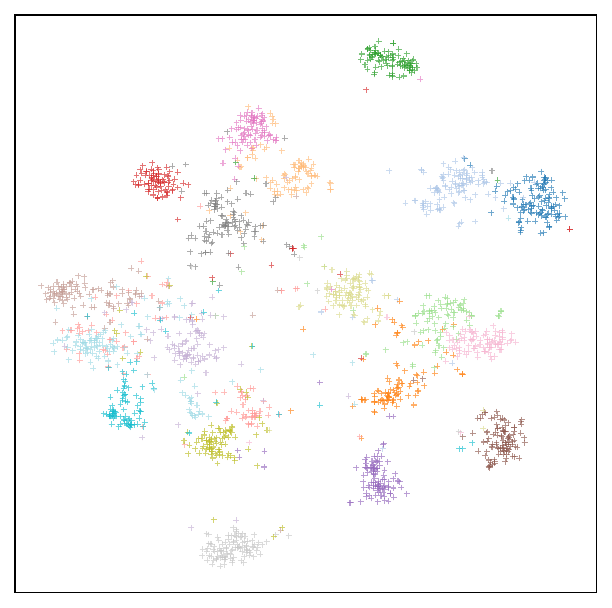} &
\includegraphics[width=0.18\textwidth]{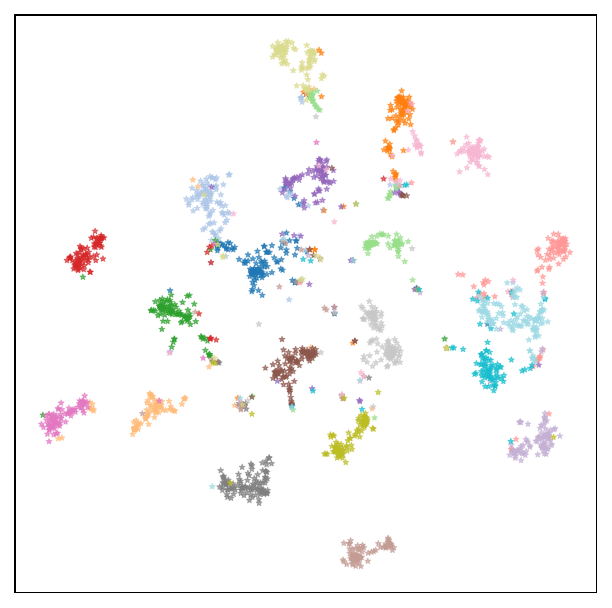} &
\includegraphics[width=0.18\textwidth]{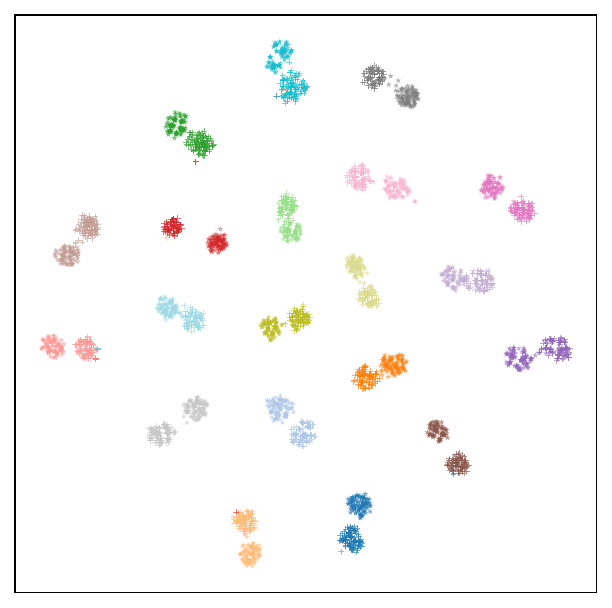} &
\includegraphics[width=0.18\textwidth]{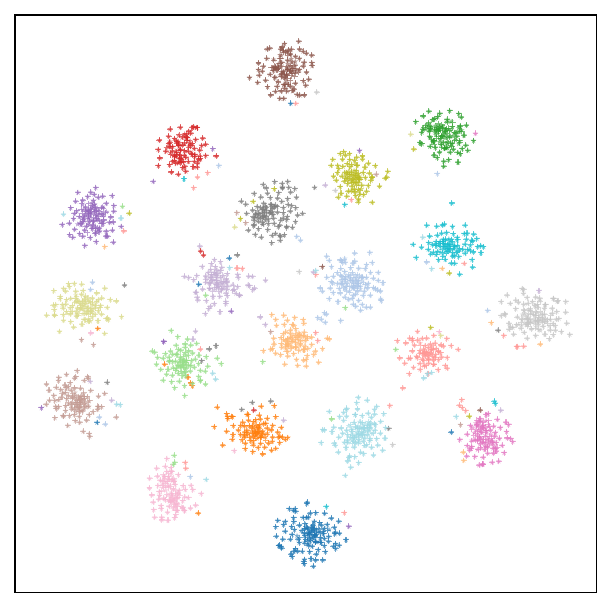} &
\includegraphics[width=0.18\textwidth]{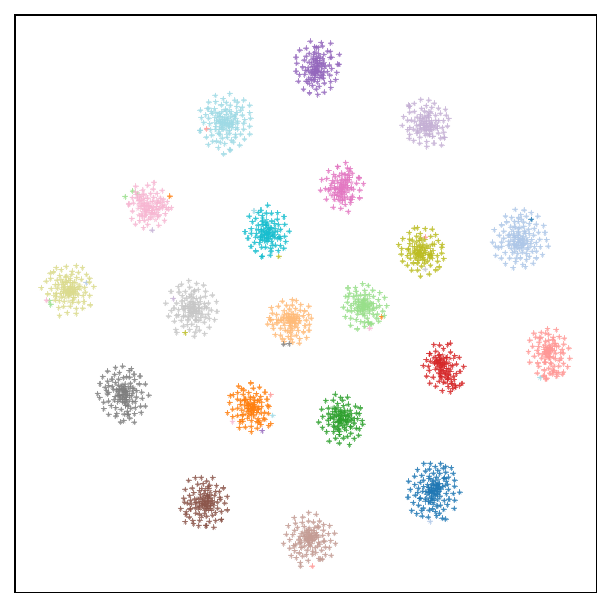} \\
{\footnotesize (a) Raw Image} &
{\footnotesize (b) Raw Text} &
{\footnotesize (c) EGICH} &
{\footnotesize (d) RSHNL} &
{\footnotesize (e) SPIRN}
\end{tabular}
}
\caption{t-SNE visualizations learned from the training data on the XMedia dataset under the medium missing scene with a 0.6 label noise ratio, where different colors denote different semantic categories.}
\label{fig:tsne}
\end{figure*}

\subsubsection{Computational Cost Analysis}
Fig.~\ref{fig:time_memory} shows the average training time and GPU memory consumption over four label-noise ratios on datasets with different modality-completeness settings. Overall, SPIRN does not introduce excessive time or memory overhead compared with existing incomplete cross-modal retrieval methods. Notably, the PGLC and CSI modules are activated only during training to facilitate robust representation learning and missing-modality imputation. During inference, SPIRN performs retrieval solely by extracting common representations using the trained encoders; therefore, it requires no additional PGLC or CSI computation beyond standard feature extraction.

\subsubsection{Prior Construction}
To isolate the effect of semantic-prior construction, we develop four variants of SPIRN. SPIRN-M uses class prototypes initialized by the mean feature of samples assigned to each class. SPIRN-E updates the prototypes using an exponential moving average, while SPIRN-L treats the prototypes as learnable parameters. SPIRN-R employs fixed random L2-normalized prototypes without explicitly enforcing orthogonality. All variants use the same network architecture, optimization strategy, and loss functions, differing only in the construction or update mechanism of the semantic priors. The complete SPIRN model uses fixed orthogonal semantic priors.
Tab.~\ref{tab:prior_construction} shows that the complete SPIRN achieves the best performance across all noise ratios, indicating that fixed orthogonal priors provide a consistent advantage over the alternative prototype constructions. Unlike data-dependent or learnable prototypes, fixed priors are not directly affected by corrupted observations. During early training, QRL identifies relatively clean samples and aligns their representations with the priors, while the orthogonal structure provides well-separated semantic directions for different categories. This consistent guidance supports robust semantic representation learning, reliable noisy-label calibration, and selective missing-modality imputation under severe data corruption. Consequently, SPIRN achieves consistently strong retrieval performance under the joint challenges of modality incompleteness and noisy labels.

\subsubsection{Visualization}
\label{sec:visualization}
To intuitively demonstrate the superiority of our proposed SPIRN, we leverage the t-SNE technique to visualize the learned feature distributions on the XMedia dataset under the 0.6 label noise ratio and the medium missing scenario. As illustrated in Fig.~\ref{fig:tsne}, we compare the visualization results of SPIRN against raw features, RSHNL, and EGICH. 
Specifically, features learned by EGICH struggle to capture consistent cross-modal semantic representations for the same category, whereas RSHNL fails to establish distinct boundaries between semantic classes, resulting in noticeable inter-class overlap and misaligned instances. In sharp contrast, our SPIRN framework yields significantly more compact and discriminative embeddings, effectively enhancing both intra-class compactness and inter-class separability. These observations further validate the robustness of SPIRN in mitigating the adverse effects of the coupled degradations.

\end{document}